\documentclass{article}
\usepackage{ijcai25}

\usepackage{times}
\usepackage{soul}
\usepackage{url}
\usepackage[hidelinks]{hyperref}
\usepackage[utf8]{inputenc}
\usepackage[small]{caption}
\usepackage{graphicx}
\usepackage{amsmath}
\usepackage{amsthm}
\usepackage{booktabs}
\usepackage{algorithm}
\usepackage{algorithmic}
\usepackage[switch]{lineno}
\usepackage{amsfonts}
\usepackage{booktabs}
\usepackage{multirow}
\usepackage{tabularx}
\usepackage{longtable}
\usepackage{colortbl}
\usepackage{diagbox}
\usepackage[noadjust]{cite}

\title{Community-Aware Temporal Walks: Parameter-Free Representation Learning on Continuous-Time Dynamic Graphs}

\author{
He Yu$^1$
\and
Jing Liu$^2$
\affiliations
$^1,^2$School of Artificial Intelligence, Xidian University, 2 South Taibai Road, Xi’an, Shaanxi 710071, China\\
$^1,^2$Guangzhou Institute of Technology, Xidian University, Knowledge City, Guangzhou, Guangdong 510555, China\\
\emails
$^1$24398862@qq.com,
$^2$neouma@mail.xidian.edu.cn	
}

\begin{document}

\maketitle

\begin{abstract}
Dynamic graph representation learning plays a crucial role in understanding evolving behaviors. However, existing methods often struggle with flexibility, adaptability, and the preservation of temporal and structural dynamics. To address these issues, we propose Community-aware Temporal Walks (CTWalks), a novel framework for representation learning on continuous-time dynamic graphs. CTWalks integrates three key components: a community-based parameter-free temporal walk sampling mechanism, an anonymization strategy enriched with community labels, and an encoding process that leverages continuous temporal dynamics modeled via ordinary differential equations (ODEs).  This design enables precise modeling of both intra- and inter-community interactions, offering a fine-grained representation of evolving temporal patterns in continuous-time dynamic graphs. CTWalks theoretically overcomes locality bias in walks and establishes its connection to matrix factorization. Experiments on benchmark datasets demonstrate that CTWalks outperforms established methods in temporal link prediction tasks, achieving higher accuracy while maintaining robustness. The implementation of our proposed method is publicly available at \url{https://github.com/leonyuhe/CTWalks}.
\end{abstract}

\section{Introduction}

Continuous-Time Dynamic Graphs (CTDGs) \cite{rossi2020temporal,xu2020inductive,wang2021causal,souza2022provably} provide a comprehensive framework for modeling temporal interactions in real-world systems. By representing entities and their time-stamped interactions as nodes and edges, CTDGs capture the evolving dynamics of diverse systems, including social networks, biological processes, knowledge graphs, and recommendation platforms \cite{nickel2015review,zhang2020gnnrec,wang2019kgat,zhang2021graph,fout2017protein}. Among the myriad tasks supported by CTDGs, \textit{temporal link prediction} \cite{luo2022neighborhood,yu2023towards} —predicting future interactions based on historical data—stands out as a fundamental problem. Success in this task not only enhances our understanding of dynamic systems but also drives practical applications such as personalized recommendations and anomaly detection. Despite significant advancements in dynamic graph representation learning \cite{kazemi2020representation,yang2023dynamic,zhu2022encoder}, existing methods face three critical challenges that limit their flexibility, adaptability, and expressiveness.

\begin{figure}[h]
    \centering
    \includegraphics[width=0.8\linewidth]{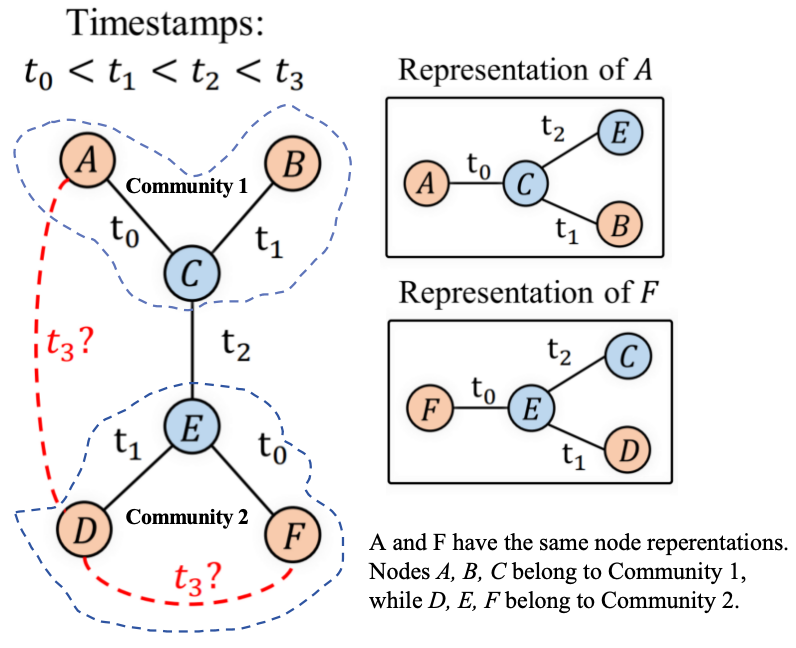}
    \caption{Traditional anonymous encodings fail to differentiate nodes with similar local structures but distinct community roles, such as $A$ and $F$. Without community context, the identical structures of $A$ and $F$ make it impossible to predict $D$’s relationship with either node.}
    \label{fig:community_representation}
\end{figure}

First, current approaches rely heavily on heuristic sampling strategies to extract key graph features, including node relationships, edge interactions, and temporal patterns. However, these strategies often demand intensive hyperparameter tuning to achieve a balance between exploration and exploitation \cite{jin2023motif,xia2021motif}. Such reliance complicates the training process and introduces sensitivity to dataset-specific characteristics, limiting their generalizability.

Second, many existing methods fail to account for mesoscopic structures, such as communities \cite{fortunato2010community,newman2004finding,Liu2014Multiobjective,Teng2021Overlapping}, which bridge local (node-level) and global (graph-level) patterns. Communities are crucial in real-world networks for distinguishing nodes with similar local structures but differing roles within their communities. Without incorporating community-aware representations, traditional anonymous encodings struggle to differentiate such nodes, resulting in ambiguity when predicting interactions, as demonstrated in Figure~\ref{fig:community_representation}.

Finally, existing methods struggle to handle the continuous nature of temporal dynamics in real-world networks. Conventional approaches often discretize time into fixed intervals or aggregate observations, which obscures important temporal information, particularly in irregularly-sampled data. Recent advances in continuous-time models, such as Neural Ordinary Differential Equations (Neural ODEs)\cite{chen2018neural}, demonstrate the potential to preserve temporal continuity by defining hidden states at all time points through differential equations. However, integrating such continuous-time methodologies with dynamic graph structures remains an active area of research\cite{jin2022mtgode, asikis2020neural, chen2023signed, sdgnn2022}, particularly in scenarios that require capturing both intra-community and inter-community temporal dependencies.

To tackle these challenges, we propose \textbf{Community-Aware Temporal Walks (CTWalks)}, a novel, parameter-free, and community-aware framework for representation learning on CTDGs. By leveraging the interplay between intra-community and inter-community dynamics, CTWalks achieves robust scalability, expressiveness, and generalizability. Contributions of this work include:

\emph{1. Community Detection on CTDGs:}
CTWalks introduces an innovative method to perform community detection on CTDGs by transforming them into a \emph{weighted temporal graph}. This transformation aggregates interactions into a static graph, where edge weights reflect interaction frequencies, thereby enabling the identification of node community labels reflecting temporal information.

\emph{2. Parameter-Free Temporal Sampling:}
CTWalks eliminates the need for handcrafted hyperparameters in sampling by adopting a community-driven mechanism. Temporal walks are adaptively performed within or across communities, encoding both structural dependencies and temporal dynamics. In the anonymization process, we embed community labels to restore contextual distinctions between nodes, reducing ambiguity and enhancing inductive capabilities.

\emph{3. Continuous-Time Dynamics via ODEs:} To explicitly model temporal dependencies and capture essential dynamic laws, CTWalks employs a Neural ODE-based process. This approach learns spatiotemporal dynamics continuously, aligning modeled dynamics with irregularly sampled observations and providing a robust basis for downstream tasks.

 \emph{4. Empirical Validation:} Extensive experiments on benchmark datasets demonstrate CTWalks’ superior performance in the temporal link prediction task. Compared to several well-known methods, CTWalks achieves higher accuracy while maintaining competitive computational efficiency.

\section{Related Work}

\emph{Dynamic graphs} \cite{kazemi2020representation,yang2023dynamic,zhu2022encoder,Ma2024Higher} provide a powerful framework for modeling systems that evolve over time, effectively capturing temporal interactions among nodes, edges, and their attributes. Depending on the representation of temporal information, dynamic graphs are broadly classified into Discrete-Time Dynamic Graphs (DTDGs) \cite{cong2021dynamic, goyal2020dyngraph2vec, pareja2020evolvegcn, sankar2020dysat, you2022roland} and Continuous-Time Dynamic Graphs (CTDGs) \cite{rossi2020temporal,xu2020inductive,wang2021causal,souza2022provably}. DTDGs approximate temporal evolution using discrete graph snapshots, which provide a coarse-grained view of dynamic systems. In contrast, CTDGs represent interactions as a sequence of time-stamped events, which allows for high temporal fidelity, enabling the representation of systems with irregular or rapid interactions. Existing representation learning methods on dynamic graphs can be categorized into two primary paradigms: \emph{message-passing-based methods} and \emph{sequence-model-based methods}. Additionally, methods based on \emph{Continuous-Time Differential Equations (CTDEs)} are discussed as a 
separate modeling approach \cite{chen2018neural,rubanova2019latent,xhonneux2020continuous}.

\emph{Message-Passing-Based Methods} extend graph neural networks (GNNs) \cite{scarselli2008graph,Mo2025AutoSGNN} to incorporate temporal information by propagating and aggregating features from neighboring nodes. For example, TGN \cite{rossi2020temporal} integrates a memory module to store historical interactions, dynamically updating it through attention-based message passing. While effective in capturing temporal dependencies, TGN suffers from high computational overhead due to frequent memory updates and neighbor aggregation, limiting its scalability for large graphs. Other approaches, such as MTSN \cite{jin2023motif}, aim to preserve structural motifs during message passing to enhance the capture of local patterns. However, the computational cost of motif construction and repeated matrix operations makes these methods impractical for large-scale applications. 

\emph{Sequence-Model-Based Methods} rely on recurrent neural networks (RNNs) \cite{elman1990finding} or related architectures to capture temporal dependencies by sequentially updating node embeddings. For instance, JODIE \cite{kumar2019predicting} employs coupled RNNs to jointly model static and dynamic embeddings, enabling tasks such as temporal link prediction. However, JODIE's reliance on hyperparameter tuning, such as decay functions for time intervals, reduces its generalizability across datasets. Similarly, CAWs \cite{wang2021causal} leverage causal anonymous walks to encode temporal sequences while preserving the temporal order of interactions. 

\emph{Continuous-Time Differential Equation (CTDE)-based Methods} \cite{chen2018neural} provide a robust framework for modeling temporal dynamics in graphs by treating temporal interactions as continuous processes. For example, ODE-RNN \cite{rubanova2019latent} leverages ODE solvers to capture the dynamics between observations, effectively bridging the gaps in irregularly sampled data. CGNN \cite{xhonneux2020continuous} introduces continuous message-passing inspired by diffusion processes, addressing challenges like over-smoothing while enabling long-range dependency modeling. TGAT \cite{xu2020inductive} employs temporal attention mechanisms coupled with Fourier-based time encoding to model temporal dependencies, showcasing the potential of attention in capturing fine-grained temporal patterns.

\section{Preliminaries}
\paragraph{Definition 1: Continuous-Time Dynamic Graph.}
A \textit{continuous-time dynamic graph} \( \mathcal{G} \) is represented as a sequence of non-decreasing chronological interactions:
\scriptsize
\begin{equation}
\mathcal{G} = \{((u_1, v_1), t_1), ((u_2, v_2), t_2), \dots\}, \quad 0 \leq t_1 \leq t_2 \leq \cdots.
\end{equation}
\normalsize
Here, each pair \((u_i, v_i)\) represents an undirected link between nodes \( u_i \) and \( v_i \), with a corresponding timestamp \( t_i \). 

\paragraph{Definition 2: Temporal Link Prediction.}
\textit{Temporal link prediction} is the task of predicting whether an interaction between two nodes will occur at a specific time, based on historical interactions in the graph. Given the set of historical interactions before time \( t \):
\begin{equation}
\mathcal{H}(t) = \{((u', v'), t') \mid t' < t\},
\end{equation}
and two nodes \( u \) and \( v \), the goal is to predict the likelihood of an interaction between \( u \) and \( v \) at time \( t \). Formally, this is expressed as:
\begin{equation}
\text{Predict:} \; P(((u, v), t) \mid \mathcal{H}(t)),
\end{equation}
where \( P \) represents the probability of the interaction, conditioned on the historical interaction sequence.

\paragraph{Definition 3: Temporal Walk.}
A \textit{temporal walk} on a CTDG \( \mathcal{G} \) is a sequence of node-time pairs:
\begin{equation}
W = \{(w_1, t_1), (w_2, t_2), \ldots, (w_l, t_l)\},
\end{equation}
where:
\begin{itemize}
    \item \( w_1 = u \) and \( t_1 = t \): \( W \) is rooted at node \( u \) at time \( t \).
    \item \( t_1 > t_2 > \ldots > t_l \): Timestamps are strictly decreasing.
    \item \( ((w_{i-1}, w_i), t_i)  \): Each step corresponds to a temporal edge in \( \mathcal{G} \).
\end{itemize}

\paragraph{Definition 4: Community.} 
A \textit{community} in a graph is a subset of nodes \( C \subseteq V \) such that nodes within \( C \) are densely connected to each other, while connections between nodes in \( C \) and those outside \( C \) are sparse. 

Formally, community detection algorithms aim to partition the node set \( V \) into \( k \) disjoint subsets \( \mathcal{C} = \{C_1, C_2, \ldots, C_k\} \), where:
\begin{equation}
C_i \cap C_j = \emptyset, \quad \forall i \neq j, \quad \text{and} \quad \bigcup_{i=1}^k C_i = V.
\end{equation}

The quality of a community partition is often quantified using metrics such as modularity \cite{newman2004finding}. Communities represent mesoscopic structures in graphs, bridging the gap between local node-level interactions and global graph-level patterns.

\section{The Proposed Method: CTWalks}
CTWalks comprises three key components: sampling, anonymization, and encoding. \emph{The complete notation system used throughout CTWalks is detailed in Appendix~A, while the algorithm's computational complexity is analyzed in Appendix~C.3.}
\subsection{Sampling}

\paragraph*{Weighted Temporal Graph Construction.} 
We propose an innovative method for community detection in CTDGs by introducing the concept of a weighted temporal graph \( G_w \), which is derived from the input CTDG \( \mathcal{G} \). The set of nodes in \( G_w \) includes all nodes appeared in \( \mathcal{G} \), and the edge weights \( w_{uv} \) in \( G_w \) encode the frequency of interactions between nodes \( u \) and \( v \), defined as:
\begin{equation}
w_{uv} = \sum_{(u, v, t) \in \mathcal{G}} \mathbb{I}(u,v),
\end{equation}
where \( \mathbb{I} \) is an indicator function counting occurrences of \( (u, v) \) in \( \mathcal{G} \). This transformation aggregates temporal interactions into a static graph while retaining the structural essence of the CTDG, thereby enabling efficient processing and community detection. By applying modularity optimization algorithms, such as Louvain \cite{newman2004fast,blondel2008fast}, on \( G_w \), the graph is partitioned into \( k \) communities \( \mathcal{C} = \{C_1, C_2, \ldots, C_k\} \).

\paragraph*{Sampling Strategy.}
To effectively capture the temporal and structural patterns in \( \mathcal{G} \), nodes are categorized based on their roles within the community structure:
\begin{itemize}
    \item \textbf{Non-Bridging Nodes}: Nodes confined within a single community \( C_i \). Their temporal walks are restricted to the neighbors within the intra-community subgraph \( G_{C_i} \).
    \item \textbf{Bridging Nodes}: Nodes that connect multiple communities. Their temporal walks are restricted to the neighbors within the inter-community subgraph \( G_I \), ensuring transitions only occur between bridging nodes.
\end{itemize}

At each step of a temporal walk, the next node \( u \) is sampled based on its temporal relationship with the current node \( v \) under the constraint \( t' < t \), where \( t \) is the current timestamp. The transition probability is defined as:
\begin{equation}
\label{eq:sampling}
P(u, t' | v, t) = \frac{e^{-(t - t')}}{\sum_{u' \in \text{N}_{\text{valid}}(v)} e^{-(t - t')}},
\end{equation}
where:
\begin{itemize}
    \item \( \text{N}_{\text{valid}}(v) \) represents the valid neighbor set of \( v \), determined by its type:
        \begin{itemize}
            \item For non-bridging nodes \( v \in V_{C_i} \), \( \text{N}_{\text{valid}}(v) = \text{N}(v) \cap G_{C_i} \), where \( \text{N}(v) \) is the neighbor set of \( v \).
            \item For bridging nodes \( v \in V_I \), \( \text{N}_{\text{valid}}(v) = \text{N}(v) \cap G_I \).
        \end{itemize}
    \item \( t' \) is the timestamp associated with the interaction between \( v \) and \( u \).
\end{itemize}

\begin{algorithm}[H]
\caption{Sampling Strategy}
\label{alg:ctwalks_sampling}
\textbf{Input:} Intra-community subgraphs \( \{G_{C_1}, G_{C_2}, \ldots, G_{C_k}\} \), inter-community subgraph \( G_I \), walk length \( l \), number of walks per node \( R \).\\
\textbf{Output:} Set of temporal walks \( \mathcal{W} \).

\begin{algorithmic}[1]
\STATE Initialize \( \mathcal{W} \gets \emptyset \)
\FOR{each node \( v \in V \)}
    \FOR{each walk \( r = 1 \) to \( R \)}
        \STATE Initialize walk \( W \gets [(v, t_1)] \), timestamp \( t \gets t_1 \)
        \IF{\( v \in V_I \)}
            \STATE \( G_{\text{current}} \gets G_I \)
        \ELSE
            \STATE Find \( C_i \) such that \( v \in V_{C_i} \)
            \STATE \( G_{\text{current}} \gets G_{C_i} \)
        \ENDIF
        \FOR{step \( i = 1 \) to \( l-1 \)}
            \STATE Identify valid neighbors \( \text{N}(v) \) with \( t' < t \)
            \IF{\( \text{N}(v) = \emptyset \)}
                \STATE Break
            \ENDIF
            \STATE Sample next node \( u \) using \( P(u, t' | v, t) \) in Eq. (\ref{eq:sampling})
            \STATE Update \( W \gets W \cup (u, t') \), \( t \gets t' \)
        \ENDFOR
        \STATE \( \mathcal{W} \gets \mathcal{W} \cup \{W\} \)
    \ENDFOR
\ENDFOR
\STATE \textbf{Return} \( \mathcal{W} \)
\end{algorithmic}
\end{algorithm}

By utilizing community partitioning to guide walk strategies, CTWalks eliminates the need for additional parameters to control walk directions, simplifying implementation and improving adaptability to diverse graph structures and temporal dynamics. Non-bridging nodes focus on intra-community walks, capturing localized patterns, while bridging nodes explore inter-community relationships. The complete CTWalks sampling process is detailed in Algorithm~\ref{alg:ctwalks_sampling}.

\subsection{Anonymization}

CTWalks replaces node identities with position-based representations while embedding both directionality and community context. For a temporal interaction \(((u, v), t_i)\), the anonymization process operates on nodes appearing in temporal walk sets originating from the source node \( u \) and target node \( v \).

\paragraph*{Community- and Direction-Aware Representation.}
The anonymized representation for a node \( w \), based on temporal walk sets from the source node \( u \) and target node \( v \), integrates both directionality and the community labels of the root nodes. This is defined as:
\scriptsize
\begin{equation}
A(w; \mathcal{S}_u, \mathcal{S}_v, C_u, C_v) = \big[A(w; \mathcal{S}_u) || C_u || A(w; \mathcal{S}_v) || C_v\big],
\end{equation}
\normalsize
where:
\begin{itemize}
    \item \( \mathcal{S}_u \) and \( \mathcal{S}_v \) are the sets of temporal walks originating from \( u \) and \( v \), respectively.
    \item \( A(w; \mathcal{S}_u) \) and \( A(w; \mathcal{S}_v) \) aggregate \( w \)'s position-based occurrence information across all walks in \( \mathcal{S}_u \) and \( \mathcal{S}_v \), respectively.
    \item \( C_u \) and \( C_v \) are the community labels of the root nodes \( u \) and \( v \), directly appended to their respective anonymization vectors.
\end{itemize}

\paragraph*{Anonymized Walk Construction.}
For a single temporal walk \( \mathcal{W} = \{w_1, w_2, \dots, w_l\} \) with ascending timestamps \( t_1 < t_2 < \dots < t_l \), the anonymized walk representation is defined as:
\begin{equation}
\mathcal{W}_{\text{anon}} = \{(A(w_i), t_i) \mid i = 1, 2, \dots, l\},
\end{equation}
where \( A(w_i) \) is the community- and direction-aware anonymized representation of node \( w_i \), incorporating information from the sets of temporal walks \( \mathcal{S}_u \) and \( \mathcal{S}_v \), as well as the community labels \( C_u \) and \( C_v \). \emph{The anonymization process is detailed in Appendix G.}

\subsection{Encoding}

The encoding mechanism in CTWalks processes anonymized temporal walks \( \mathcal{W}_{\text{anon}} \) by combining \emph{continuous temporal evolution} and \emph{instantaneous updates}, while integrating \emph{community-aware information}. This ensures the final state \( h_l \) captures temporal, structural, and community-aware information for downstream tasks.

The process begins by initializing the cumulative hidden state \( h'_0 = \mathbf{0} \). Before the first step, the instantaneous hidden state \( h_1 \) is computed as:
\begin{equation}
h_1 = g(h'_0, A(w_1)).
\end{equation}

For each subsequent step \( i \) (\( 1 \leq i < l \)), the encoding alternates between:
\begin{enumerate}
    \item \textbf{Continuous Integration:} Update \( h'_i \) by integrating the temporal evolution function \( f(h, t) \) over \([t_{i}, t_{i+1}]\):
    \begin{equation}
    h'_i = \int_{t_{i}}^{t_{i+1}} f(h_{i}, t) \, dt.
    \end{equation}
    \item \textbf{Instantaneous Update:} Compute \( h_{i+1} \), incorporating \( h'_i \) and \( A(w_{i+1}) \):
    \begin{equation}
    \label{eq:instant}
    h_{i+1} = g(h'_i, A(w_{i+1})).
    \end{equation}
\end{enumerate}
We define \( g \) and \( f \) as parameterized models, \( g \) updates instantaneous states, while \( f \) governs continuous temporal integration. \emph{Details are provided in Appendix~C.1.}

\begin{algorithm}[H]
\caption{Encoding of CTWalks}
\label{alg:ctwalks_encoding}
\textbf{Input:} Anonymized temporal walk \( \mathcal{W}_{\text{anon}} = \{(A(w_1), t_1), \ldots, (A(w_l), t_l)\} \).\\
\textbf{Output:} Final walk representation \( h_l \).

\begin{algorithmic}[1]
\STATE Initialize \( h'_0 \gets \mathbf{0} \)
\STATE Compute \( h_1 \gets g(h'_0, A(w_1)) \) 
\FOR{\( i = 1 \) to \( l-1 \)}
    \STATE Update \( h'_i \) via continuous integration
    \STATE Compute \( h_{i+1} \) via instantaneous update
\ENDFOR
\STATE \textbf{Return} \( h_l \)
\end{algorithmic}
\end{algorithm}

\begin{figure}[H]
    \centering
    \includegraphics[width=0.8\linewidth]{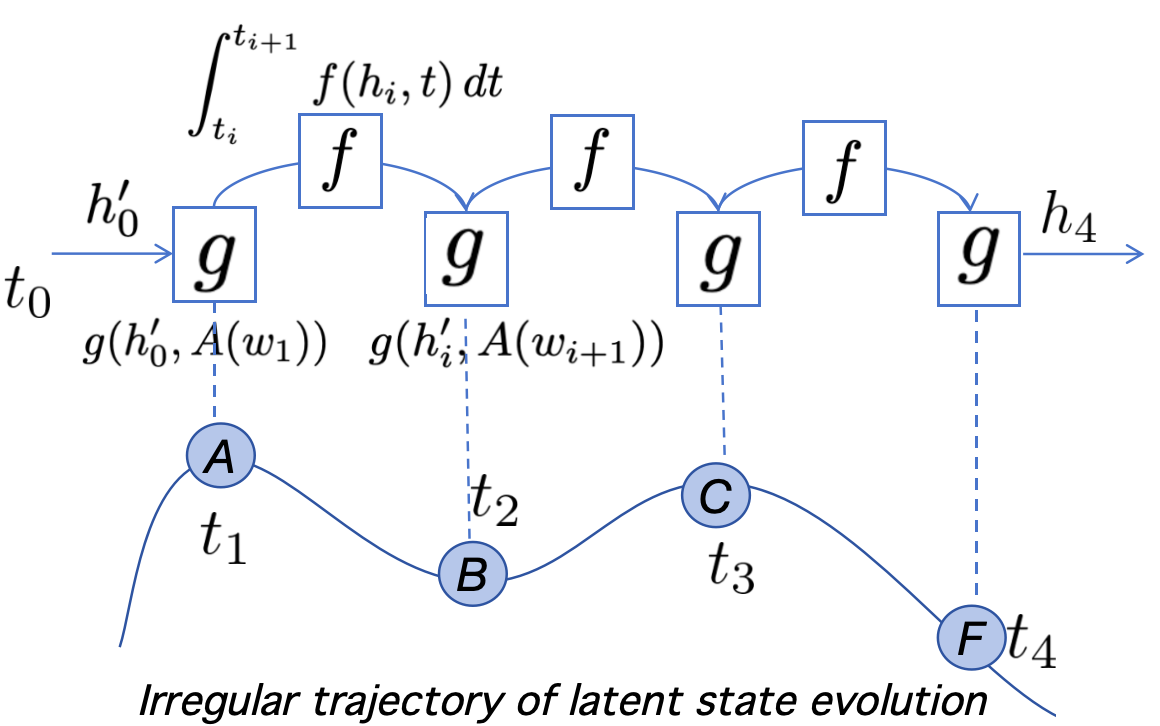}
    \caption{Illustration of the encoding mechanism in CTWalks. Each node in the temporal walk is processed through instantaneous updates (\( g \)) and continuous temporal integration (\( f \)) along an irregular time trajectory. The final hidden state \( h_4 \) represents the walk encoding, incorporating both structural and community-aware information.}
    \label{fig:ctwalks_encoding}
\end{figure}

By alternating between cumulative updates and instantaneous hidden states, the encoding ensures \( h_l \) is temporally coherent, structurally rich, and embeds community context for robust downstream tasks. The detailed implementation of this process is provided in Algorithm~\ref{alg:ctwalks_encoding}, and an illustration is shown in Fig.~\ref{fig:ctwalks_encoding}. \emph{For a detailed discussion, please refer to Appendix H.}

\subsection{Extension for Attributed Graphs}

To encode node and edge attributes, the encoding process can be extended by modifying the instantaneous update function \( g \) as follows:
\begin{equation}
    h_i = g(h'_{i-1}, A(w_i) || X_{w_i} || X_{e_i}),
\end{equation}

where:
\begin{itemize}
    \item \( X_{w_i} \) represents the attributes of node \( w_i \),
    \item \( X_{e_i} \) represents the attributes of edge \( e_i = \{w_{i-1}, w_i\} \),
    \item \( || \) denotes the concatenation operation.
\end{itemize}

In this extended formulation, node attributes \( X_{w_i} \) and edge attributes \( X_{e_i} \) introduce additional context for real-world attributed dynamic graphs.
Several methods \cite{yang2013community, yang2013overlapping, du2017community} focus on community detection in attributed graphs, leveraging node attributes to improve accuracy. This extension enhances the model's flexibility and expressiveness, enabling it to capture not only the structural and temporal information but also the rich attributes associated with nodes and edges.

\section{Theoretical Analysis}

Dynamic graph sampling inherently involves both spatial structures and temporal sequences. To isolate the impact of temporal sequences and focus on spatial sampling, it is essential to establish an unbiased spatial sampling environment as a foundation. To this end, we analyze the potential biases introduced by traditional random walks on unweighted, undirected static graphs, which may limit their ability to comprehensively explore the graph structure. Furthermore, we examine whether a parameter-free walk strategy, guided by community partitions and distinguishing intra-community and inter-community transitions, can effectively address these biases.

By simplifying the problem to static graphs, we aim to eliminate the confounding effects of temporal dynamics and concentrate on evaluating the locality bias and the theoretical advantages of community-guided walks in an idealized spatial context.

\subsection*{Analysis 1: Overcoming Locality Bias}

Let a traditional random walk generate a sequence \( \{w_0, w_1, ..., w_{L_w-1}\} \), where \( w_t \in V \). At step \( t+1 \), the walk transitions to a neighbor of \( w_t \) with probability \( 1/d(w_t) \), where \( d(w_t) \) is the degree of node \( w_t \). Let \( P_{uv} \) denote the one-step transition probability from node \( u \) to \( v \), defined as:
\begin{equation}
P_{uv}= 
\begin{cases}
1/d(u),& \text{if } (u,v) \in E, \\
0,              & \text{otherwise.}
\end{cases}
\end{equation}

Suppose \( w_0 = u \) and \( w_t = v \), where \( u, v \in V \) and \( w_0, w_1, ..., w_{t-1} \neq v \). The probability that the walk starts at \( u \) and visits \( v \) for the first time at time \( t \) is given by:
\begin{equation}
r^t_{uv} = \sum_{j \in \mathcal{N}(u)} P_{uj} \cdot r^{t-1}_{jv} = \frac{1}{d(u)} \sum_{j \in \mathcal{N}(u)} r^{t-1}_{jv},
\end{equation}
where \( \mathcal{N}(u) \) is the set of neighbors of node \( u \). This equation shows that \( r^t_{uv} \) is the mean of the \((t-1)\)-step probabilities \( r^{t-1}_{jv} \) over all neighbors of \( u \). The reliance on all neighbors dilutes the transition probability towards distant nodes, reinforcing locality bias.

\textbf{Lemma 1.} In CTWalks, let nodes \( u \) and \( v \) belong to two different communities, and \( u, v \in V_c \) are bridging nodes. The probability that the walk starts from \( u \) and visits \( v \) for the first time at time \( t \) satisfies:
\begin{equation}
r^t_{uv} \geq \frac{1}{d(u)} \sum_{j \in \mathcal{N}(u)} r^{t-1}_{jv},
\end{equation}
with equality when all neighbors of \( u \) are bridging nodes (\( \mathcal{N}(u) = \mathcal{N}_{\text{inter}}(u) \)).

\textbf{Proof.} Partition \( \mathcal{N}(u) \) into two disjoint subsets:
\begin{itemize}
    \item \( \mathcal{N}_{\text{inter}}(u) = \mathcal{N}(u) \cap V_c \), the set of inter-community neighbors (bridging nodes),
    \item \( \mathcal{N}_{\text{intra}}(u) = \mathcal{N}(u) - \mathcal{N}_{\text{inter}}(u) \), the set of intra-community neighbors.
\end{itemize}

The total probability of transitioning to neighbors of \( u \) can be expressed as:
\begin{equation}
\sum_{j \in \mathcal{N}(u)} r^{t-1}_{jv} = \sum_{j \in \mathcal{N}_{\text{inter}}(u)} r^{t-1}_{jv} + \sum_{j \in \mathcal{N}_{\text{intra}}(u)} r^{t-1}_{jv}.
\end{equation}

In CTWalk, intra-community transitions cannot reach nodes in different communities. Thus, for any \( j \in \mathcal{N}_{\text{intra}}(u) \), \( r^{t-1}_{jv} = 0 \). The total probability simplifies to:
\begin{equation}
\sum_{j \in \mathcal{N}(u)} r^{t-1}_{jv} = \sum_{j \in \mathcal{N}_{\text{inter}}(u)} r^{t-1}_{jv}.
\end{equation}

The transition probability \( r^t_{uv} \) from \( u \) to \( v \) in CTWalks is:
\begin{equation}
\label{r_uv_t}
r^t_{uv} = \frac{1}{|\mathcal{N}_{\text{inter}}(u)|} \sum_{j \in \mathcal{N}_{\text{inter}}(u)} r^{t-1}_{jv}.
\end{equation}

Since \( |\mathcal{N}_{\text{inter}}(u)| \leq d(u) \), it follows that:
\begin{equation}
\frac{1}{|\mathcal{N}_{\text{inter}}(u)|} \geq \frac{1}{d(u)}.
\end{equation}

Substituting this into the Eq.(\ref{r_uv_t}) for \( r^t_{uv} \), we obtain:
\begin{equation}
r^t_{uv} = \frac{1}{|\mathcal{N}_{\text{inter}}(u)|} \sum_{j \in \mathcal{N}_{\text{inter}}(u)} r^{t-1}_{jv} \geq \frac{1}{d(u)} \sum_{j \in \mathcal{N}(u)} r^{t-1}_{jv}.
\end{equation}

Equality holds when all neighbors of \( u \) are bridging nodes (\( \mathcal{N}(u) = \mathcal{N}_{\text{inter}}(u) \)).

\textbf{Lemma 1} demonstrates that CTWalks reduces locality bias by prioritizing inter-community transitions for bridging nodes. Unlike traditional random walks, which distribute transition probabilities evenly among all neighbors, CTWalks separates intra- and inter-community contributions, enabling effective exploration of global structures. 

\emph{We also theoretically establish the connection between CTWalks and matrix factorization, details are provided in Appendix F.}

\section{Experiments}
\subsection{Experimental Setting}

\paragraph{Baselines and Datasets.}
CTWalks is evaluated against six state-of-the-art baselines for continuous-time dynamic graphs, including message passing-based methods (DyRep \cite{trivedi2019dyrep}, TGAT \cite{xu2020inductive}, and TGN \cite{rossi2020temporal}) and sequential model-based methods (CTDNE \cite{nguyen2018continuous}, JODIE \cite{kumar2019predicting}, and CAWs \cite{wang2021causal}) .  The evaluation is conducted on five real-world datasets spanning diverse domains: social networks, email communications, e-commerce, online education, and student forums. For instance, UCI \cite{leskovec2016snap} tracks social interactions in a university network; Enron \cite{leskovec2016snap} captures email exchanges among employees; Taobao \cite{zhu2018learning} records user-item interactions with encoded action types such as clicks and purchases; MOOC \cite{leskovec2016snap} logs student activities on educational platforms; and Wikipedia \cite{kumar2019predicting} is a bipartite interaction network consisting of editor-page and user-post activities. The dataset statistics are summarized in Table~\ref{tab:dataset}. \emph{Additional details on the baselines and datasets are available in the Appendix~D}.

\begin{table}[h]
    \centering
    \small
    \resizebox{0.85\linewidth}{!}{%
    \begin{tabular}{l|ccccc}
    \hline
    \textbf{Dataset} & \textbf{Nodes} & \textbf{Temporal Edges} & \textbf{Duration (seconds)} & \textbf{Interaction Intensity} \\ \hline
    UCI & 1,899 & 59,835 & 16,621,303 & $3.79 \times 10^{-6}$ \\
    MOOC & 7,144 & 411,749 & 2,572,086 & $4.48 \times 10^{-5}$ \\
    Enron & 143 & 62,617 & 72,932,520 & $6.50 \times 10^{-5}$ \\
    Taobao & 64,703 & 77,436 & 36,000 & $6.64 \times 10^{-5}$ \\
    Wikipedia & 9,227 & 157,474 & 2,678,373 & $1.27 \times 10^{-5}$ \\
    \hline
    \end{tabular}%
    }
    \caption{Dataset statistics. Interaction intensity is calculated as $2|E|/(|V|T)$, where $T$ is the total time range in seconds, $|V|$ and $|E|$ are number of nodes and temporal links.}
    \label{tab:dataset}
\end{table}

\paragraph{Data Preparation.}

The data preparation process involves three key steps. First, all edges in the input temporal graph are sorted by their timestamps to ensure chronological consistency. This step facilitates training on past edges while testing on future edges. Next, the dataset is split into training, validation, and testing sets, adhering to a chronological ratio of 70\% (training), 15\% (validation), and 15\% (testing). Finally, negative sampling is performed by generating edges absent in the original graph to form negative edge sets, ensuring a balanced dataset. Both positive and negative samples are used to train, validate, and test the model. Based on the training data, a Weighted Temporal Graph is constructed to capture the aggregate interaction strength between nodes over time. For nodes in the validation and testing datasets that are absent in the Weighted Temporal Graph:
\begin{itemize}
    \item \textbf{Non-bridging nodes:} If all neighbors of the node belong to the same community, the node is assigned to that community.
    \item \textbf{Bridging nodes:} If the node’s neighbors belong to multiple communities, the node is assigned to a community with a probability proportional to the sum of the edge weights connecting the node to that community, defined as:
    \begin{equation}
        P(C_i | v) = \frac{\sum_{u \in N(v) \cap C_i} w_{vu}}{\sum_{u \in N(v)} w_{vu}},
    \end{equation}
    where \( P(C_i | v) \) is the probability of node \( v \) belonging to community \( C_i \), \( N(v) \) represents the neighbors of \( v \), and \( w_{vu} \) is the edge weight between \( v \) and \( u \).
\end{itemize}

\paragraph{Evaluation Tasks.} 
We assess the performance of CTWalks and baselines on \emph{link prediction} tasks, which are further categorized into \textit{transductive} and \textit{inductive} settings: In the \emph{transductive link prediction} task, temporal links between all nodes are used for training up to a specific time point, with the remaining links allocated for validation and testing. In the \emph{inductive link prediction} task, the model is evaluated on its ability to predict links involving nodes unseen during training. Two specific scenarios are considered: 
1. \textit{New-Old}: Interactions between unseen (new) and observed (old) nodes. 2. \textit{New-New}: Interactions exclusively between unseen nodes. To construct these scenarios, 10\% of the nodes are masked during training, and interactions associated with them are removed. Validation and testing sets are limited to interactions involving these masked nodes.
\begin{figure}[H]
    \centering
    \includegraphics[width=0.8\linewidth]{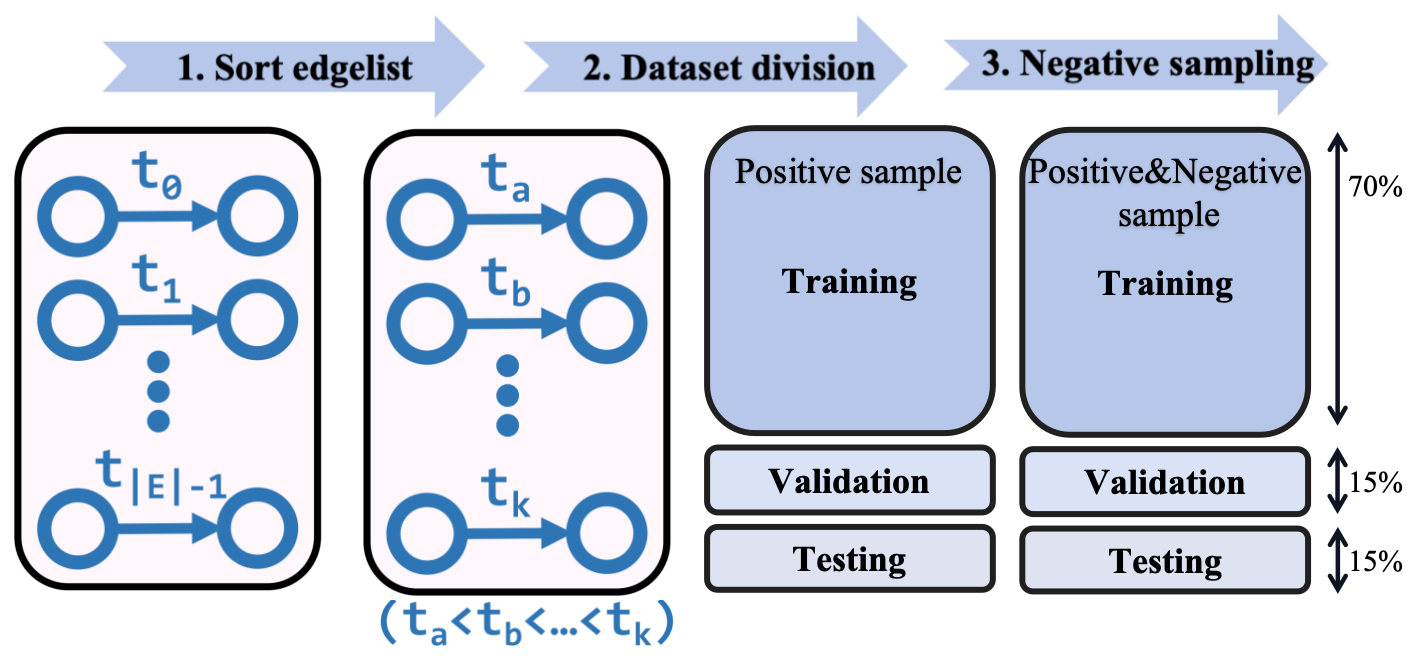}
    \caption{Illustration of the data preparation process for link prediction. The process includes sorting edges by timestamp, splitting into training, validation, and testing sets, and performing negative sampling to ensure balanced datasets.}
    \label{fig:data_preparation}
\end{figure}

\paragraph{Training Details.} 
For evaluation, we report results using two widely adopted metrics: Area Under the ROC Curve (AUC) and Average Precision (AP) \emph{which are detailed in Appendix D}. These metrics effectively capture model performance across both transductive and inductive tasks. To ensure robustness, each experiment is repeated five times with different random seeds, and we report the average performance.

\begin{table}[t]
\label{tab:result}
    \centering
    \resizebox{\columnwidth}{!}{%
    \begin{tabular}{c|c|l|lllll}
    \hline
    \multicolumn{2}{c|}{Task} & Methods & \multicolumn{1}{c}{UCI} &  \multicolumn{1}{c}{MOOC} & \multicolumn{1}{c}{Enron} & \multicolumn{1}{c}{Taobao} & \multicolumn{1}{c}{Wikipedia}\\
    \hline
    \multirow{16}{*}{\rotatebox[origin=c]{90}{Inductive}}
    &\multirow{8}{*}{\rotatebox[origin=c]{90}{new v.s. new}}
    & DyRep& 63.76 $\pm$ 4.67&  74.07$\pm$ 1.88& 72.74$\pm$ 0.44& 87.62 $\pm$ 1.23$^\dagger$& 67.07 $\pm$ 1.26\\
    && TGAT& 76.33 $\pm$ 1.38&  70.04 $\pm$ 1.35& 63.34 $\pm$ 1.82& 76.99 $\pm$ 1.54& 90.77 $\pm$ 0.95\\
    && TGN& 79.37 $\pm$ 1.49&  77.07 $\pm$ 1.88& 80.92 $\pm$ 0.02& 84.74 $\pm$ 0.44& 63.76 $\pm$ 4.67\\
    && CTDNE& 62.93 $\pm$ 0.72&   72.01 $\pm$ 0.29& 66.49 $\pm$ 1.21& 78.01 $\pm$ 0.08& 62.35 $\pm$ 1.10\\
    && JODIE& 70.31 $\pm$ 0.53&  75.31 $\pm$ 4.14& 70.95 $\pm$ 0.85& 81.53 $\pm$ 2.12& 72.65 $\pm$ 0.63\\
    && CAWs& 88.12$\pm$ 0.88$^\dagger$&  90.10 $\pm$ 0.35$^\dagger$& \textbf{96.27 $\pm$ 1.16}& 86.34 $\pm$ 2.95& \textbf{96.36 $\pm$ 1.48}\\
    && \textbf{CTWalks}& \textbf{90.71 $\pm$ 0.08}&   \textbf{92.86 $\pm$ 0.56}& 95.20 $\pm$ 0.68$^\dagger$& \textbf{96.98 $\pm$ 0.41}& 94.16 $\pm$ 0.25$^\dagger$\\
   
    \cline{2-8}
    &\multirow{8}{*}{\rotatebox[origin=c]{90}{new v.s. old}}
    & DyRep& 93.28 $\pm$ 3.01$^\dagger$&  88.23$\pm$ 1.35& \textbf{94.39 $\pm$ 0.59}& 89.36 $\pm$ 1.48$^\dagger$& 76.72 $\pm$ 1.36\\
    && TGAT& 76.33 $\pm$ 1.03&  67.40 $\pm$ 1.71& 61.80 $\pm$ 1.31& 65.24 $\pm$ 0.98& 95.47$\pm$ 1.18\\
    && TGN& 87.13 $\pm$ 1.07&  72.23 $\pm$ 1.20$^\dagger$& 77.98 $\pm$ 0.45& 88.39 $\pm$ 0.32& 90.28 $\pm$ 0.96\\
    && CTDNE& 71.11 $\pm$ 0.74&  69.10 $\pm$ 0.88& 64.66 $\pm$ 0.41 & 68.33 $\pm$ 0.05& 88.39 $\pm$ 1.08\\
    && JODIE& 71.61 $\pm$ 0.37&  88.20 $\pm$ 1.92& 85.73 $\pm$ 1.36& 80.53 $\pm$ 2.13& 74.78 $\pm$ 0.22\\
    && CAWs& 91.25 $\pm$ 0.18&  90.30 $\pm$ 0.08$^\dagger$& 93.22 $\pm$ 1.28$^\dagger$& 88.76 $\pm$ 1.18& \textbf{96.19 $\pm$ 0.88}\\
     && \textbf{CTWalks}& \textbf{95.70 $\pm$ 0.17}&  \textbf{92.08 $\pm$ 0.20}& 92.52 $\pm$ 0.21& \textbf{93.98 $\pm$ 0.44}& 95.15 $\pm$ 0.24$^\dagger$\\
    \hline
    
    \multicolumn{2}{c|}{\multirow{8}{*}{\rotatebox[origin=c]{90}{Transductive}}}
    & DyRep& 95.23 $\pm$ 1.48$^\dagger$& 90.49 $\pm$ 0.24& \textbf{96.71 $\pm$ 0.80}& 83.11 $\pm$ 1.13& 77.40 $\pm$ 1.79\\
    \multicolumn{2}{c|}{}& TGAT& 77.67 $\pm$ 1.02& 72.09 $\pm$ 1.51& 60.88 $\pm$ 1.34& 62.88 $\pm$ 1.46& 96.36 $\pm$ 1.21$^\dagger$\\
    \multicolumn{2}{c|}{}& TGN& 83.36 $\pm$ 1.23& 73.09 $\pm$ 0.03& 68.12 $\pm$ 1.21& 87.71 $\pm$ 0.04$^\dagger$& 95.23 $\pm$ 0.25\\
    \multicolumn{2}{c|}{}& CTDNE& 76.89 $\pm$ 0.92& 73.03 $\pm$ 0.32& 89.36 $\pm$ 0.73& 65.08 $\pm$ 0.02& 92.43 $\pm$ 0.27\\
    \multicolumn{2}{c|}{}& JODIE& 74.63 $\pm$ 0.52& 90.50 $\pm$ 0.85& 60.36 $\pm$ 0.65 & 82.02 $\pm$ 0.31& 89.30 $\pm$ 0.22\\
    \multicolumn{2}{c|}{}& CAWs& 95.25 $\pm$ 0.06& 94.28 $\pm$ 0.29$^\dagger$& 91.64 $\pm$ 0.55& 85.88 $\pm$ 0.37& \textbf{98.67 $\pm$ 0.27}\\
    \multicolumn{2}{c|}{}& \textbf{CTWalks}& \textbf{98.05 $\pm$ 0.09}&  \textbf{94.33 $\pm$ 0.08}& 92.54 $\pm$ 0.58$^\dagger$& \textbf{92.06 $\pm$ 0.16}&   95.14 $\pm$ 0.25\\
    \hline
    \end{tabular}
    }
    
    \vspace{-1.5mm}
    \caption{Transductive and inductive link prediction performances w.r.t. AUC. We use \textbf{bold font }and $^\dagger$ to highlight the best and second best performances.}
    \label{tab:auc results}
    \vspace{-2.0mm}
\label{tab: results}
\end{table}

\subsection{Experimental Results and Discussion}

We report the AUC performance of CTWalks and six state-of-the-art baselines across inductive and transductive link prediction tasks on various datasets. Table~\ref{tab: results} summarizes the results, highlighting the effectiveness of CTWalks in both inductive and transductive settings.

In the inductive setting, CTWalks achieves remarkable results, particularly for the challenging "new vs. new" links. On most of datasets, CTWalks significantly outperforms the best-performing baseline CAWs, with improvements of up to 5.14\% on the UCI dataset and 2.16\% on Taobao. Similarly, in the "new vs. old" scenario, CTWalks consistently surpasses the baselines, achieving an AUC of 95.70 on UCI and 92.08 on MOOC, compared to the respective baseline bests of 93.28 by DyRep and 90.30 by CAWs. These results highlight CTWalks' superior generalization ability, particularly for interactions involving previously unseen nodes. 

In the transductive setting, CTWalks again demonstrates strong performance, achieving the best or second-best results across all datasets. On the Taobao dataset, CTWalks attains an AUC of 92.06. Similarly, on the MOOC dataset, CTWalks achieves an AUC of 94.33, showcasing its competitive edge even on attributed datasets.

The success of CTWalks can be attributed to its ability to integrate temporal dynamics and structural information effectively, leveraging its community-aware temporal walk encoding to capture intricate graph patterns. Its superior performance across diverse datasets highlights its generalizability and robustness, setting a benchmark for future research in dynamic graph learning.

\subsection{Ablation Study}

We conducted an ablation study to evaluate the contributions of key components in CTWalks, with results presented in Table~\ref{tab:ctwalks_ablation}. The study involved systematically removing essential mechanisms to understand their impact on the performance of CTWalks.

\begin{table}[t]
\scriptsize \centering
\resizebox{\columnwidth}{!}{%
\begin{tabular}{cllll}
\hline
\textbf{No.} & \textbf{Ablation} & \multicolumn{1}{c}{\textbf{UCI}} & \multicolumn{1}{c}{\textbf{Taobao}}  & \multicolumn{1}{c}{\textbf{MOOC}} \\ \hline
1. & original method         & \textbf{92.94 $\pm$ 0.21} & \textbf{94.88 $\pm$ 0.36} & \textbf{92.07 $\pm$ 0.21} \\
2. & remove intra-community walk & \textbf{89.28 $\pm$ 0.66} & \textbf{88.45 $\pm$ 0.41} & \textbf{85.69 $\pm$ 0.31} \\
3. & remove inter-community walk & \textbf{86.90  $\pm$ 0.17} & \textbf{87.91 $\pm$ 0.39} &  \textbf{86.25 $\pm$ 1.99} \\
4. & remove community walk & \textbf{85.94 $\pm$ 0.85} & \textbf{86.01 $\pm$ 0.12} & \textbf{85.12 $\pm$ 0.20} \\
5. & remove anonymous community label & \textbf{88.94 $\pm$ 0.61} & \textbf{87.01 $\pm$ 0.14} & \textbf{86.12 $\pm$ 0.30} \\
6. & remove continuous evolution & \textbf{79.38 $\pm$ 2.05} & \textbf{84.44 $\pm$ 0.56} & \textbf{83.72 $\pm$ 0.44} \\ \hline  
\end{tabular}}
\caption{\small Ablation study on CTWalks in terms of AUC scores on \textit{all inductive} test links.}
\label{tab:ctwalks_ablation}
\end{table}

The first three experiments assess the influence of the community-aware sampling mechanism. When the restrictions on intra-community and inter-community walks are removed, nodes are allowed to perform unrestricted temporal walks, with the selection of the next node determined by time-biased probabilities that prioritize temporally closer events. The resulting performance degradation across these variants underscores the significance of preserving community boundaries. This highlights the effectiveness of the proposed community-aware sampling strategy in capturing both local and global temporal dynamics. To further validate the advantages of community walks in enhancing network embedding learning, we conducted experiments on purely static graphs and compared against classical graph node embedding algorithms. \emph{Detailed analyses are provided in Appendix~E.}

The fourth experiment removes the use of community labels during the anonymized walk encoding. This change prevents the model from distinguishing between structurally similar walks based on their community context, reducing its ability to effectively generalize across different network structures.

Finally by removing the continuous integration step, the cumulative hidden state \( h'_i \) is no longer updated through integration over temporal intervals. Instead, the encoding process relies solely on the instantaneous update function Eq.(\ref{eq:instant}),
where the hidden state at each step is directly computed without accounting for continuous temporal evolution. This simplification significantly impacts performance, particularly on datasets with sparse interactions like UCI, as it eliminates the ability to capture nuanced spatiotemporal dynamics over time intervals, underscoring the critical role of the continuous integration mechanism.

These results validate that each component contributes significantly to the overall performance of CTWalks, with the interplay of community-aware sampling and continuous evolution mechanisms being particularly crucial.

\section{Discussion and Conclusion}

The effectiveness of our model relies heavily on modularity-based community detection methods, such as Louvain, which ties its performance to the quality of the initial community partitions. In networks with overlapping or poorly-defined communities, this dependency may hinder the efficacy of the community-aware sampling mechanism. To address this, future work could explore adaptive sampling techniques that dynamically refine community boundaries or integrate multi-scale community detection methods to enhance robustness. 

Additionally, incorporating continuous temporal dynamics via ODE solvers poses scalability challenges for extremely large graphs. To address these issues, approximate solutions such as logarithmic normalization of time intervals and parallel batch processing (\emph{see Appendix C.2}) have been employed to reduce computational overhead. Nonetheless, future research should prioritize the development of lightweight integration techniques or hybrid approaches that strike a balance between computational efficiency and modeling fidelity.

CTWalks represents a community-aware, parameter-free framework for representation learning on CTDGs. Unlike existing methods, CTWalks simultaneously incorporates intra- and inter-community dynamics, enabling the effective modeling of mesoscopic structures within graphs. Our contributions span several dimensions: (1) a novel temporal walk sampling strategy that adaptively captures community-driven dynamics without requiring extensive parameter tuning, (2) an anonymized encoding process augmented with community labels for generalization, and (3) a continuous temporal integration mechanism to capture nuanced spatiotemporal dependencies. Experimental results on diverse datasets consistently demonstrate the superiority of CTWalks in both inductive and transductive tasks.

This work lays the groundwork for future research in dynamic graph learning. Directions for future exploration include designing scalable methods for handling larger and more complex CTDGs, developing robust community detection mechanisms, and addressing challenges associated with data bias and ethical use in real-world applications. With its potential to generalize across domains, CTWalks represents a significant step forward in understanding and modeling the evolving dynamics of complex systems.

\bibliographystyle{ieeetr}

\newpage
\appendix
\onecolumn
\section{Notations and Definitions}
\addcontentsline{toc}{section}{Appendix A: Notations and Definitions}

\begin{table}[H]
    \centering
    \renewcommand{\arraystretch}{1.3}
    \begin{tabular}{|c|l|}
        \hline
        \textbf{Symbol} & \textbf{Definition} \\ \hline
        \( \mathcal{G} = \{(e_i, t_i)\}_{i=1}^N \) & A continuous-time dynamic graph (CTDG) with time-stamped interactions. \\ \hline
        \( G_w = (V, E_w) \) & The weighted temporal graph with aggregated interaction weights. \\ \hline
        \( w_{uv} \) & Weight of the edge \((u, v)\), representing the total number of interactions. \\ \hline
        \( \mathcal{C} = \{C_1, C_2, \ldots, C_k\} \) & Partitioned communities in the graph using modularity-based algorithms. \\ \hline
        \( G_{C_i} = (V_{C_i}, E_{C_i}) \) & Subgraph corresponding to community \( C_i \). \\ \hline
        \( G_I = (V_I, E_I) \) & Inter-community subgraph comprising bridging nodes and edges. \\ \hline
        \( V_I \) & Set of bridging nodes connecting multiple communities. \\ \hline
        \( \text{N}(v) \) & Neighbor set of node \( v \). \\ \hline
        \( W \) & A temporal walk defined as \( \{(w_1, t_1), (w_2, t_2), \ldots, (w_l, t_l)\} \). \\ \hline
        \( P_{\text{intra}}(u | v, t) \) & Transition probability for intra-community walks based on temporal constraints. \\ \hline
        \( P_{\text{inter}}(u | v, t) \) & Transition probability for inter-community walks based on temporal constraints. \\ \hline
        \( A(w; \mathcal{S}_u, \mathcal{S}_v, C_u, C_v) \) & Anonymization operation with community label integration. \\ \hline
        \( \mathcal{W}_{\text{anon}} \) & Anonymized temporal walk incorporating community-aware representations. \\ \hline
        \( h_i \) & Instantaneous hidden state at step \( i \) of the walk. \\ \hline
        \( h'_i \) & Cumulative hidden state at step \( i \) updated through continuous integration. \\ \hline
        \( g(h'_{i-1}, A(w_i) \) & Instantaneous update function for hidden states. \\ \hline
        \( f(h, t) \) & Temporal evolution function for continuous integration. \\ \hline
        \( \int_{t_i}^{t_{i+1}} f(h_i, t) \, dt \) & Continuous integration over the temporal interval \([t_i, t_{i+1}]\). \\ \hline
        \( h_l \) & Final representation of the temporal walk \( W \). \\ \hline
        \( \lambda = 2|E| / (|V|T) \) & Average link stream intensity, measuring interaction density over time. \\ \hline
    \end{tabular}
    \caption{Key Notations and Definitions for CTWalks}
    \label{tab:notations}
\end{table}

\noindent
\textbf{Additional Notes:}
\begin{itemize}
    \item The definitions aim to maintain clarity for each symbol and its application within the CTWalks framework.
    \item Symbols like \( g(h'_{i-1}, A(w_i)) \) and \( f(h, t) \) are parameterized models, where \( g \) handles discrete updates and \( f \) continuous temporal dynamics.
    \item The temporal walks \( W \) integrate both intra-community and inter-community dynamics, with anonymization enhancing generalization across unseen nodes.
    \item Community-aware sampling eliminates the need for additional parameters to control local/global walks, enhancing model simplicity and robustness.
\end{itemize}

\noindent

\section{Community Detection Algorithms}
\label{sec:community_detection}

\subsection{Louvain Method}

The Louvain method \cite{newman2004fast,blondel2008fast} is an efficient modularity optimization algorithm widely used for community detection in undirected weighted graphs. It operates in two iterative phases:

\begin{enumerate}
    \item \textbf{Local Optimization Phase}: Initially, each node is treated as a separate community. The algorithm iteratively moves nodes between neighboring communities to maximize the modularity gain \( \Delta Q \), defined as:
    \begin{equation}
    \Delta Q = \frac{\sum_{\text{in}} + 2k_{\text{in}}}{2m} - \left(\frac{\sum_{\text{tot}} + k_{\text{tot}}}{2m}\right)^2 - \left(\frac{\sum_{\text{tot}}}{2m}\right)^2,
    \end{equation}
    where \( \sum_{\text{in}} \) is the sum of edge weights within a community, \( k_{\text{in}} \) and \( k_{\text{tot}} \) are the total weights of edges connected to the node inside the community and in the entire graph, respectively, and \( m \) is the total edge weight in the graph.
    
    \item \textbf{Community Aggregation Phase}: Once no further modularity improvement is possible, each community is treated as a supernode, forming a new graph. The process is repeated until the modularity \( Q \) no longer improves significantly.
\end{enumerate}

The Louvain method is highly scalable and effective, making it suitable for large-scale graphs.

\subsection{Other Community Detection Algorithms}

Several alternative community detection methods exist, each with unique strengths and weaknesses:

\begin{itemize}
    \item \textbf{Spectral Clustering}: Based on the eigenvectors of the graph Laplacian, this method identifies \( k \) communities.
    \begin{itemize}
        \item \textit{Advantages}: Effective for small-scale graphs and capable of capturing complex community structures.
        \item \textit{Disadvantages}: Computationally expensive for large graphs.
    \end{itemize}

    \item \textbf{Random Walk-Based Methods (e.g., Infomap)}: These minimize the description length of random walks to identify communities.
    \begin{itemize}
        \item \textit{Advantages}: Naturally handle weighted graphs and hierarchical community structures.
        \item \textit{Disadvantages}: High computational resource requirements.
    \end{itemize}

    \item \textbf{Label Propagation}: Communities are formed through the iterative propagation and convergence of node labels.
    \begin{itemize}
        \item \textit{Advantages}: Simple and fast, well-suited for dynamic graphs.
        \item \textit{Disadvantages}: Susceptible to local optima and instability.
    \end{itemize}

    \item \textbf{Modularity Maximization}: A family of algorithms that optimize modularity \( Q \). The Louvain method is a prominent example.
    \item \textbf{Deep Learning-Based Methods}: Algorithms like DeepWalk and Node2Vec learn node embeddings to capture community structures.
    \begin{itemize}
        \item \textit{Advantages}: Integrate node attributes and graph topology information.
        \item \textit{Disadvantages}: Require significant computational resources and are sensitive to hyperparameters.
    \end{itemize}
\end{itemize}

In our framework, the Louvain method is employed for community detection due to its efficiency and suitability for weighted graphs. This method provides a solid foundation for subsequent node community classification and boundary node detection. Future work may explore advanced algorithms tailored to dynamic weighted graphs for more accurate and robust community detection.

\section{Batching and Complexity Analysis}

\subsection{Function Definitions: \(g\) and \(f\)}

In the CTWalks framework, two key functions, \(g\) and \(f\), are defined to model discrete updates and continuous temporal evolution, respectively. Their complementary roles enable precise representation of both instantaneous and time-dependent dynamics within continuous-time dynamic graphs.

\paragraph{1. Instantaneous Update Function \(g\)} 

The function \(g\) handles the \textbf{discrete, step-wise state updates} at specific nodes along a temporal walk. For each node \(w_i\), the hidden state \(h_i\) is computed as:
\begin{equation}
h_i = g(h'_{i-1}, A(w_i)),
\end{equation}
where:
\begin{itemize}
    \item \(h'_{i-1}\) is the cumulative hidden state from the previous step.
    \item \(A(w_i)\) is the anonymized representation of node \(w_i\).
    \item \(g\) is instantiated as a Gated Recurrent Unit (GRU) \cite{cho2014learning}, designed to capture both structural and temporal information.
\end{itemize}

The GRU-based formulation of \(g\) is given by:
\begin{equation}
z_t = \sigma(W_z h_{t-1} + U_z x_t + b_z),
\end{equation}
\begin{equation}
r_t = \sigma(W_r h_{t-1} + U_r x_t + b_r),
\end{equation}
\begin{equation}
\tilde{h}_t = \tanh(W_h (r_t \odot h_{t-1}) + U_h x_t + b_h),
\end{equation}
\begin{equation}
h_t = z_t \odot \tilde{h}_t + (1 - z_t) \odot h_{t-1},
\end{equation}
where:
\begin{itemize}
    \item \(z_t\), \(r_t\), and \(\tilde{h}_t\) are the update gate, reset gate, and candidate hidden state, respectively.
    \item \(x_t = A(w_i)\) represents the input node features.
    \item \(\sigma(\cdot)\) and \(\tanh(\cdot)\) denote the sigmoid and hyperbolic tangent activation functions, respectively.
    \item \(\odot\) represents element-wise multiplication.
\end{itemize}

\paragraph{2. Continuous Temporal Evolution Function \(f\)}

The function \(f\) models the \textbf{continuous evolution of the hidden state} over time intervals between nodes. Unlike \(g\), which incorporates node-specific input features, \(f\) focuses solely on temporal dynamics and acts on the output of \(g\). The cumulative hidden state \(h'_i\) at step \(i\) is computed by integrating \(f\) over the time interval \([t_i, t_{i+1}]\):
\begin{equation}
h'_i = \int_{t_i}^{t_{i+1}} f(h_i, t) \, dt,
\end{equation}
where:
\begin{itemize}
    \item \(h_i\) is the instantaneous hidden state computed by \(g\),
    \item \(t_i\) and \(t_{i+1}\) are the timestamps associated with the current and next nodes in the temporal walk,
    \item \(f(h, t)\) is parameterized as a GRU-like model, similar to \(g\), but without external inputs \(x_t\).
\end{itemize}

The formulation of \(f\) is as follows:
\begin{equation}
z_t = \sigma(W_z h_{t-1} + b_z),
\end{equation}
\begin{equation}
r_t = \sigma(W_r h_{t-1} + b_r),
\end{equation}
\begin{equation}
\tilde{h}_t = \tanh(W_h (r_t \odot h_{t-1}) + b_h),
\end{equation}
\begin{equation}
h_t = z_t \odot \tilde{h}_t + (1 - z_t) \odot h_{t-1},
\end{equation}
where:
\begin{itemize}
    \item \(z_t\), \(r_t\), and \(\tilde{h}_t\) are the update gate, reset gate, and candidate hidden state, respectively,
    \item \(W_z\), \(W_r\), and \(W_h\) are weight matrices,
    \item \(b_z\), \(b_r\), and \(b_h\) are biases,
    \item \(\sigma(\cdot)\) and \(\tanh(\cdot)\) denote the sigmoid and hyperbolic tangent activation functions.
\end{itemize}

By removing external inputs \(x_t\), \(f\) focuses purely on modeling the temporal continuity of the hidden state, allowing it to integrate spatiotemporal dependencies over time intervals.

Together, \(g\) and \(f\) enable CTWalks to seamlessly integrate spatial and temporal dynamics, delivering a robust framework for learning representations on continuous-time dynamic graphs.

\subsection{Batching Analysis}
Efficiently encoding temporal walks poses a significant challenge due to the irregular timestamps and varying lengths of walks. For \(C\) temporal walks, each consisting of \(l\) steps, solving \(C \times l\) independent ordinary differential equations (ODEs) is computationally expensive. To address this, we employ a \textbf{time reparameterization strategy}, unifying the integration intervals of all walks into a standard range \([0, 1]\). This unification allows batch processing of the walks, enabling significant computational efficiency\cite{chen2020neural}.

\paragraph{Reparameterization for Unified Time Intervals}
Consider the \(c\)-th walk with timestamps \([t_{\text{start}}^c, t_{\text{end}}^c]\), where \(t_{\text{start}}^c\) is the time of the first node in the walk, and \(t_{\text{end}}^c\) is the time of the last node. We reparameterize the time \(t\) within this interval using a new variable \(s \in [0, 1]\), defined as:
\begin{equation}
    s = \frac{t - t_{\text{start}}^c}{t_{\text{end}}^c - t_{\text{start}}^c}, \quad \text{thus,} \quad t = s \cdot (t_{\text{end}}^c - t_{\text{start}}^c) + t_{\text{start}}^c.
\end{equation}

Under this transformation, the instantaneous hidden state \(h_t\) at time \(t\) is reformulated as \(\tilde{h}_s\), such that:
\begin{equation}
    \tilde{h}_s = h_{t} = h_{s \cdot (t_{\text{end}}^c - t_{\text{start}}^c) + t_{\text{start}}^c}.
\end{equation}
This allows all walks, regardless of their original time intervals, to operate within a common temporal scale of \([0, 1]\).

\paragraph{Reformulated ODE for Batch Processing}
The cumulative hidden state for the \(c\)-th walk is computed by solving an ODE over the interval \([t_{\text{start}}^c, t_{\text{end}}^c]\). After reparameterization, the ODE is reformulated for \(\tilde{h}_s\) as follows:
\begin{equation}
    \frac{d\tilde{h}_s}{ds} = f(h_t, t) \cdot \frac{dt}{ds} = f(\tilde{h}_s, s \cdot (t_{\text{end}}^c - t_{\text{start}}^c) + t_{\text{start}}^c) \cdot (t_{\text{end}}^c - t_{\text{start}}^c),
\end{equation}
where \(\frac{dt}{ds} = t_{\text{end}}^c - t_{\text{start}}^c\).

The initial and final states are transformed as:
\begin{equation}
    \tilde{h}_0 = h_{t_{\text{start}}^c}, \quad \tilde{h}_1 = h_{t_{\text{end}}^c}.
\end{equation}

The solution for the cumulative hidden state at the end of the walk becomes:
\begin{equation}
    h_{t_{\text{end}}^c} = \text{ODESolve}(h_{t_{\text{start}}^c}, f, t_{\text{start}}^c, t_{\text{end}}^c) = \text{ODESolve}(\tilde{h}_0, \tilde{f}, 0, 1).
\end{equation}
Here, the reparameterized function \(\tilde{f}\) ensures that all walks are solved over a common interval, enabling efficient batch computation.

\paragraph{Parallelism Across Walks}
While each walk still involves step-wise computations within the reparameterized interval, the unification of time intervals to \([0, 1]\) allows a \textbf{single solver instance} to process multiple walks in parallel. This is achieved by stacking the dynamics of all walks into a joint system:
\begin{equation}
    \frac{d\tilde{h}_{\text{joint}}}{ds} = \begin{bmatrix}
        f_1(\tilde{h}_1, s \cdot \Delta t_1 + t_{\text{start}}^1) \cdot \Delta t_1 \\
        f_2(\tilde{h}_2, s \cdot \Delta t_2 + t_{\text{start}}^2) \cdot \Delta t_2 \\
        \vdots \\
        f_C(\tilde{h}_C, s \cdot \Delta t_C + t_{\text{start}}^C) \cdot \Delta t_C
    \end{bmatrix},
\end{equation}
where \(\Delta t_c = t_{\text{end}}^c - t_{\text{start}}^c\). The solver processes this joint system, reducing the computational overhead from \(C\) independent solver calls to a single batched solver call.

\paragraph{Batch Processing Efficiency}
For walks with \(l = 1\), the reparameterization directly enables batch processing since the time interval for all walks is normalized to \([0, 1]\). However, for walks with \(l > 1\), each step within the walk corresponds to a distinct interval, making true parallelism challenging. Solving multiple sub-intervals within each walk requires sequential processing.

To fully enable parallel processing for \(l > 1\), further strategies, such as normalizing intermediate step intervals or discretizing the ODE solutions, could be employed. These approaches remain an open area for further optimization.

\paragraph{Handling Large Time Intervals}
In real-world scenarios, large time intervals \(\Delta t_c\) can cause numerical instability. To mitigate this, we normalize the intervals while preserving relative temporal differences using a logarithmic transformation:
\begin{equation}
    \Delta t_{\text{scaled}} = \log_{10}(\Delta t + 1),
\end{equation}
ensuring that smaller intervals retain fine-grained dynamics while larger intervals are compressed to a manageable scale.

This batching mechanism combines \textbf{time reparameterization} and \textbf{joint ODE solving}, significantly improving computational efficiency and enabling scalable representation learning on dynamic graphs.

\subsection{Complexity Analysis}

We analyze the time complexity of the main components in CTWalks, including community detection, temporal walk sampling, anonymization, continuous evolution, and instantaneous updates. The detailed breakdown is as follows:

\paragraph{Community Detection.} 
Community detection is performed using a modularity-based algorithm, such as Louvain, which has a time complexity of \( O(|E| \log n) \), where \( |E| \) is the number of edges and \( n \) is the number of nodes. This step is executed once as preprocessing.

\paragraph{Temporal Walk Sampling.}
In each batch, \( BC \) walks are generated, where \( B \) is the batch size (number of root nodes) and \( C \) is the number of walks per root node. The sampling process involves \( l \) steps per walk, with each step querying neighbors at a complexity of \( O(k_s) \), where \( k_s \) is the average number of neighbors. The total complexity for temporal walk sampling is \( O(BCk_s) \).

\paragraph{Walk Anonymization.}
For anonymization, each walk requires \( O(l) \) operations to identify unique nodes and assign positional encodings. Across \( BC \) walks, the total complexity becomes \( O(BCl) \).

\paragraph{Continuous Evolution.}
Continuous evolution integrates the temporal evolution function over time intervals using an ODE solver, requiring \( F \) function evaluations per step. For \( BC \) walks with \( l \) steps, the total complexity is \( O(BClFd_h^2) \). Using the batch processing optimization introduced in Section B.1, this is reduced to \( O(lFd_h^2) \).

\paragraph{Instantaneous Updates.}
The instantaneous updates compute the hidden state at each step using a parameterized function, resulting in a complexity of \( O(ld_h^2) \).

\subsubsection*{Subtotal (Sampling and Anonymization).}
The combined complexity for sampling and anonymization, including temporal walk sampling and walk anonymization, is:
\begin{equation}
O(BC(k_s + l)).
\end{equation}

\subsubsection*{Subtotal (Encoding).}
The combined complexity for continuous evolution and instantaneous updates is:
\begin{equation}
O(lFd_h^2).
\end{equation}

\paragraph{Total Complexity.}
Combining all components, the total time complexity of CTWalks per epoch is:
\begin{equation}
O(|E| \log n) + O(BC(k_s + l)) + O(lFd_h^2).
\end{equation}

Table~\ref{tab:complexity_analysis} summarizes the complexity of each component.

\begin{table}[h]
    \centering

    \begin{tabular}{l|c}
        \hline
        \textbf{Component} & \textbf{Time Complexity} \\ \hline
        Community Detection & \( O(|E| \log n) \) \\
        Temporal Walk Sampling & \( O(BCk_s) \) \\
        Walk Anonymization & \( O(BCl) \) \\ \hline
        \textbf{Subtotal (Sampling and Anonymization)} & \( O(BC(k_s + l)) \) \\ \hline
        Continuous Evolution & \( O(lFd_h^2) \) \\
        Instantaneous Updates & \( O(ld_h^2) \) \\ \hline
        \textbf{Subtotal (Encoding)} & \( O(lFd_h^2) \) \\ \hline
        \textbf{Total} & \( O(|E| \log n) + O(BC(k_s + l)) + O(lFd_h^2) \) \\ \hline
    \end{tabular}
        \caption{Time Complexity Analysis of CTWalks}
        \label{tab:complexity_analysis}
\end{table}

\section{Additional Experimental Details}

\subsection{Dataset Details and Preprocessing}

We evaluate our method on five real-world datasets across diverse domains. The details are as follows:
\begin{itemize}
    \item \textbf{UCI:} Contains 1,899 nodes and 59,835 temporal edges, representing message interactions in a university social forum. No additional preprocessing is required.\footnote{Dataset available at: \url{https://snap.stanford.edu/data/CollegeMsg.html}}
    \item \textbf{MOOC:} Logs 7,145 nodes and 411,749 temporal edges, capturing student interactions with course units over one month.\footnote{Dataset available at: \url{http://snap.stanford.edu/jodie/mooc.csv}}
    \item \textbf{Enron:} Includes 184 nodes and 125,235 temporal edges, recording email communications over several years.\footnote{Dataset available at: \url{http://www.cs.cmu.edu/~enron/}}
    \item \textbf{Taobao:} Contains 987,994 nodes and 2,099,520 temporal edges, representing user-item interactions with encoded action types (e.g., clicks, purchases).\footnote{Dataset available at: \url{https://tianchi.aliyun.com/dataset/dataDetail?dataId=649&userId=1}}
    \item \textbf{Wikipedia:} Bipartite graph with 9,227 nodes and 157,474 temporal edges, representing editor-page interactions encoded with LIWC features.\footnote{Dataset available at: \url{http://snap.stanford.edu/jodie/wikipedia.csv}}
\end{itemize}

\noindent \textbf{Data Preprocessing:} The following steps ensure consistency and facilitate evaluation:
\begin{enumerate}
    \item All temporal edges are sorted chronologically to maintain temporal consistency.
    \item Edges are split into training (70\%), validation (15\%), and testing (15\%) sets.
    \item Negative sampling is performed to generate edges absent in the original graph, ensuring a balanced dataset.
\end{enumerate}

\subsection{Baselines and Hyperparameter Tuning}

\paragraph{Baseline Methods:}  
To evaluate the performance of our proposed method, we compare it with six state-of-the-art baselines specifically designed for continuous-time dynamic graphs (CTDGs):

\begin{itemize}
    \item \textbf{CTDNE}: CTDNE extends static network embedding techniques to CTDGs by leveraging temporal random walks and a skip-gram model to learn node representations. This method captures the temporal dependency of interactions.\footnote{CTDNE repository: \url{https://github.com/LogicJake/CTDNE}}
    \item \textbf{JODIE}: JODIE employs two mutually influenced recurrent neural networks (RNNs) to update the latent states of interacting nodes. The model is capable of predicting future embedding trajectories based on past interactions.\footnote{JODIE repository: \url{https://github.com/claws-lab/jodie}}
    \item \textbf{DyRep}: DyRep integrates sequence modeling with an attentive message-passing mechanism, incorporating 2-hop temporal neighborhood information to produce time-aware embeddings. The loss function is built upon temporal point processes.\footnote{DyRep repository: \url{https://github.com/uoguelph-mlrg/LDG}}
    \item \textbf{TGAT}: TGAT applies temporal encoding via random Fourier time encodings and attentively aggregates temporal neighborhood information to generate embeddings. The model supports multi-hop temporal message aggregation.\footnote{TGAT repository: \url{https://github.com/StatsDLMathsRecomSys/Inductive-representation-learning-on-temporal-graphs}}
    \item \textbf{TGN}: TGN proposes a memory-based message-passing framework, which updates node memories dynamically while combining key designs from JODIE and TGAT to enhance its capability to model temporal interactions.\footnote{TGN repository: \url{https://github.com/twitter-research/tgn}}
    \item \textbf{CAWs}: CAWs encode and aggregate anonymous temporal walks using RNNs, followed by pooling mechanisms to generate temporal node embeddings. Its community-aware design enables it to capture complex temporal dynamics.\footnote{CAWs repository: \url{https://github.com/snap-stanford/CAW}}
\end{itemize}

\paragraph{Hyperparameter Tuning:}  
To ensure a fair comparison, we carefully adapt each baseline to our unified evaluation framework. Key hyperparameters are tuned through grid search to optimize model performance. Below, we summarize the tuning strategies for each method:

\begin{itemize}
    \item \textbf{TGAT}: 
    Neighborhood Sampling Degree is tuned from \{8, 16, 32, 64\}; Model Layers are searched over \{1, 2, 3\}; Attention Heads (default product attention) are tuned from \{2, 4, 6\}; and Embedding Dimensions are tuned from \{16, 32, 64, 100\}.
    \item \textbf{JODIE}: Embedding Dimensions are tuned from \{32, 64, 128, 256\}.
    \item \textbf{DyRep}: Neighborhood Sampling Degree is tuned from \{10, 16, 32, 64\}, and Attention Layers are tuned over \{1, 2, 3\}.
    \item \textbf{TGN}: Attention Heads are tuned from \{2, 4, 8\}; Embedding Dimensions for nodes, time, and messages are tuned over \{16, 32, 64, 100\}; and Memory Dimensions are searched over \{4, 16, 32, 64, 172\}.
    \item \textbf{CAWs}: Walk Length is tuned over \{1, 2, 3\}; Number of Walks is tuned from \{16, 32, 64, 128\}; and Time Decay Factor is tuned over \{10$^{-7}$, 10$^{-6}$, 10$^{-5}$, 10$^{-4}$\}. Walk pooling is set to summation for consistency and simplicity.
    \item \textbf{CTDNE}: Skip-gram Window Size is tuned from \{3, 5, 10\}.
\end{itemize}

\begin{table}[h]
    \centering
    \small
    \resizebox{0.8\linewidth}{!}{%
    \begin{tabular}{l|ccc}
    \hline
    \textbf{Method} & \textbf{Hyperparameter} & \textbf{Search Range} & \textbf{Optimal Value} \\ \hline
    TGAT & Neighborhood Sampling Degree & \{8, 16, 32, 64\} & 32 \\
         & Model Layers & \{1, 2, 3\} & 2 \\
         & Attention Heads & \{2, 4, 6\} & 4 \\
         & Embedding Dimensions & \{32, 64, 128, 256\}& 256\\ \hline
    JODIE & Embedding Dimensions & \{32, 64, 128, 256\} & 256\\ \hline
    DyRep & Neighborhood Sampling Degree & \{10, 16, 32, 64\} & 16 \\
          & Attention Layers & \{1, 2, 3\} & 2 \\ \hline
    TGN & Attention Heads & \{2, 4, 8\} & 4 \\
        & Embedding Dimensions & \{32, 64, 128, 256\} & 256\\
        & Memory Dimensions & \{4, 16, 32, 64, 172\} & 64 \\ \hline
    CAWs & Walk Length & \{1, 2, 3\} & 3 \\
         & Number of Walks & \{16, 32, 64\} & 64 \\
         & Time Decay Factor & \{10$^{-7}$, 10$^{-6}$, 10$^{-5}$, 10$^{-4}$\} & 10$^{-6}$ \\ \hline
    CTDNE & Skip-gram Window Size & \{3, 5, 10\} & 5 \\ \hline
    \end{tabular}%
    }
    \caption{Hyperparameter search ranges and optimal values for baseline methods.}
    \label{tab:baseline_hyperparams}
\end{table}

\paragraph{Implementation Details:}  
All baselines are implemented using publicly available codebases. We ensure consistency in training and evaluation settings, including dataset splits and metrics, to enable fair comparisons. For additional implementation details, such as the specific training configurations and runtime environments, refer to Appendix~C.3.

\subsection{Implementation Details}

\paragraph{Code Availability:}  
Our implementation of CTWalks is publicly available at \url{https://github.com/leonyuhe/CTWalks}. The repository includes detailed instructions for dataset preparation, model training, and hyperparameter tuning.

\paragraph{General Training Settings:}  
CTWalks is trained across all datasets using the following settings:
\begin{itemize}
    \item Optimizer: Adam with a learning rate of $10^{-4}$ and batch size of 32.
    \item Early Stopping: Training stops if validation performance does not improve for three consecutive epochs.
    \item Epoch Limit: Maximum 50 epochs.
\end{itemize}

\paragraph{Hyperparameter Tuning:}  
To ensure optimal performance, key hyperparameters controlling walk sampling are tuned via grid search. The following summarizes the hyperparameter configurations for all datasets:
\begin{itemize}
    \item \textbf{Walk Length ($l$):} \{1, 2, 3\}
    \item \textbf{Number of Walks per Node ($C$):} \{16, 32, 64\}
    \item \textbf{ODE Solver:} Fixed-step Runge-Kutta 3/8 method with a step size of 0.125.
\end{itemize}

\noindent \textbf{Evaluation Tasks:}
\begin{itemize}
    \item \textbf{Transductive:} Models are trained on all edges up to a specific time point and tested on future edges.
    \item \textbf{Inductive:} Evaluates the ability to predict links involving unseen nodes. Two scenarios are considered:
    \begin{enumerate}
        \item \textit{New-Old:} Interactions between unseen and observed nodes.
        \item \textit{New-New:} Interactions exclusively between unseen nodes.
    \end{enumerate}
\end{itemize}

\noindent \textbf{Hardware Configuration:} Experiments are conducted on an Ubuntu Linux server equipped with four NVIDIA GeForce RTX 3090 GPUs, two Intel(R) Xeon(R) Silver 4210 CPUs (2.20GHz), and 252GB of RAM. Each experiment is repeated five times, and the average results are reported.

\noindent \textbf{Evaluation Metrics:}
\begin{itemize}
    \item AUC (Area Under the ROC Curve): Measures the quality of predictions.
    \item AP (Average Precision): Evaluates the precision-recall relationship, detail in Table~\ref{tab:ap_results}.
\end{itemize}

\begin{table}[t]
\centering
\resizebox{0.8\columnwidth}{!}{%
\begin{tabular}{c|c|l|lllll}
\hline
\multicolumn{2}{c|}{Task} & Methods & \multicolumn{1}{c}{UCI} &  \multicolumn{1}{c}{MOOC} & \multicolumn{1}{c}{Enron} & \multicolumn{1}{c}{Taobao} & \multicolumn{1}{c}{Wikipedia}\\
\hline
\multirow{16}{*}{\rotatebox[origin=c]{90}{Inductive}}
&\multirow{8}{*}{\rotatebox[origin=c]{90}{new v.s. new}}
& DyRep& 64.55 $\pm$ 4.22 & 74.93 $\pm$ 1.77 & 73.41 $\pm$ 0.41 & 88.33 $\pm$ 1.15$^\dagger$ & 67.89 $\pm$ 1.21 \\
&& TGAT& 77.20 $\pm$ 1.32 & 71.15 $\pm$ 1.27 & 64.11 $\pm$ 1.69 & 77.45 $\pm$ 1.45 & 91.32 $\pm$ 0.89 \\
&& TGN& 80.28 $\pm$ 1.39 & 78.31 $\pm$ 1.77 & 81.78 $\pm$ 0.02 & 85.34 $\pm$ 0.41 & 64.56 $\pm$ 4.33 \\
&& CTDNE& 63.75 $\pm$ 0.67 & 72.91 $\pm$ 0.27 & 67.14 $\pm$ 1.15 & 78.77 $\pm$ 0.08 & 63.09 $\pm$ 1.05 \\
&& JODIE& 71.45 $\pm$ 0.49 & 76.08 $\pm$ 4.00 & 71.66 $\pm$ 0.81 & 82.18 $\pm$ 2.04 & 73.54 $\pm$ 0.61 \\
&& CAWs& 88.78 $\pm$ 0.82$^\dagger$ & 90.89 $\pm$ 0.33$^\dagger$ & \textbf{96.95 $\pm$ 1.09} & 86.99 $\pm$ 2.80 & \textbf{96.91 $\pm$ 1.41} \\
&& \textbf{CTWalks}& \textbf{91.26 $\pm$ 0.08} & \textbf{93.72 $\pm$ 0.52} & 95.91 $\pm$ 0.62$^\dagger$ & \textbf{97.75 $\pm$ 0.38} & 94.94 $\pm$ 0.24$^\dagger$ \\
\cline{2-8}
&\multirow{8}{*}{\rotatebox[origin=c]{90}{new v.s. old}}
& DyRep& 94.11 $\pm$ 2.91$^\dagger$ & 89.07 $\pm$ 1.29 & \textbf{95.18 $\pm$ 0.55} & 90.11 $\pm$ 1.42$^\dagger$ & 77.45 $\pm$ 1.27 \\
&& TGAT& 77.10 $\pm$ 0.99 & 68.40 $\pm$ 1.62 & 62.11 $\pm$ 1.24 & 65.93 $\pm$ 0.93 & 96.12 $\pm$ 1.10 \\
&& TGN& 87.78 $\pm$ 1.01 & 73.02 $\pm$ 1.14$^\dagger$ & 78.56 $\pm$ 0.42 & 89.09 $\pm$ 0.30 & 90.96 $\pm$ 0.89 \\
&& CTDNE& 71.89 $\pm$ 0.70 & 70.10 $\pm$ 0.83 & 65.44 $\pm$ 0.38 & 69.02 $\pm$ 0.05 & 89.06 $\pm$ 1.02 \\
&& JODIE& 72.55 $\pm$ 0.35 & 89.00 $\pm$ 1.81 & 86.54 $\pm$ 1.29 & 81.23 $\pm$ 2.03 & 75.66 $\pm$ 0.20 \\
&& CAWs& 92.10 $\pm$ 0.16 & 91.14 $\pm$ 0.07$^\dagger$ & 94.10 $\pm$ 1.22$^\dagger$ & 89.56 $\pm$ 1.10 & \textbf{96.99 $\pm$ 0.81} \\
&& \textbf{CTWalks}& \textbf{96.35 $\pm$ 0.16} & \textbf{93.81 $\pm$ 0.18} & 93.17 $\pm$ 0.19 & \textbf{94.78 $\pm$ 0.41} & 96.01 $\pm$ 0.22$^\dagger$ \\
\hline
\multicolumn{2}{c|}{\multirow{8}{*}{\rotatebox[origin=c]{90}{Transductive}}}
& DyRep& 96.11 $\pm$ 1.42$^\dagger$ & 91.22 $\pm$ 0.22 & \textbf{97.41 $\pm$ 0.75} & 83.79 $\pm$ 1.07 & 78.23 $\pm$ 1.70 \\
\multicolumn{2}{c|}{}& TGAT& 78.11 $\pm$ 0.99 & 72.95 $\pm$ 1.43 & 61.32 $\pm$ 1.27 & 63.76 $\pm$ 1.38 & 96.89 $\pm$ 1.11$^\dagger$ \\
\multicolumn{2}{c|}{}& TGN& 84.01 $\pm$ 1.16 & 74.21 $\pm$ 0.03 & 69.01 $\pm$ 1.14 & 88.21 $\pm$ 0.04$^\dagger$ & 95.93 $\pm$ 0.23 \\
\multicolumn{2}{c|}{}& CTDNE& 77.55 $\pm$ 0.86 & 74.03 $\pm$ 0.29 & 90.08 $\pm$ 0.68 & 65.78 $\pm$ 0.02 & 93.11 $\pm$ 0.26 \\
\multicolumn{2}{c|}{}& JODIE& 75.29 $\pm$ 0.49 & 91.40 $\pm$ 0.78 & 61.11 $\pm$ 0.62 & 82.77 $\pm$ 0.29 & 90.44 $\pm$ 0.21 \\
\multicolumn{2}{c|}{}& CAWs& 95.97 $\pm$ 0.05 & 95.11 $\pm$ 0.27$^\dagger$ & 92.32 $\pm$ 0.51 & 86.88 $\pm$ 0.34 & \textbf{99.21 $\pm$ 0.26} \\
\multicolumn{2}{c|}{}& \textbf{CTWalks}& \textbf{98.65 $\pm$ 0.08} & \textbf{95.88 $\pm$ 0.07} & 93.35 $\pm$ 0.54$^\dagger$ & \textbf{92.98 $\pm$ 0.13} & 97.88 $\pm$ 0.23\\
\hline
\end{tabular}
}
\caption{Transductive and inductive link prediction performances w.r.t. AP. We use \textbf{bold font} and $^\dagger$ to highlight the best and second-best performances.}
\label{tab:ap_results}
\end{table}

\section{Experiment on Purely Static Graphs}
To further validate the benefits of community walks in enhancing network embedding learning, we conducted experiments on purely static graphs, focusing on the community structure without considering temporal information. Specifically, we applied our method, CTWalks, which incorporates community walks to generate node sequences, and subsequently used the word2Vec \cite{mikolov2013efficient} framework to generate embeddings from these sequences. By deliberately excluding temporal data and leveraging word2Vec, our approach aims to highlight the effectiveness of CTWalks in capturing community structures and achieving competitive performance on static graphs compared to classical graph embedding algorithms.

\subsection*{Experimental Setup}
We evaluated CTWalks (CTW) against six widely recognized baseline methods: node2vec (N2V) \cite{grover2016node2vec}, DeepWalk (DW) \cite{perozzi2014deepwalk}, LINE \cite{tang2015line}, Struc2Vec (S2V) \cite{ribeiro2017struc2vec}, GraphSage (GS) \cite{hamilton2017graphsage}, and M-NMF (MN) \cite{wang2017community}. To ensure a fair comparison, the implementations of these methods were obtained from publicly available repositories. Specifically, Node2vec, DeepWalk, and M-NMF were implemented using the Karate Club library\footnote{\url{https://github.com/benedekrozemberczki/karateclub}}. The implementation of LINE was sourced from its official repository\footnote{\url{https://github.com/tangjianpku/LINE}}. Similarly, Struc2Vec\footnote{\url{https://github.com/leoribeiro/struc2vec}} and GraphSage\footnote{\url{https://github.com/williamleif/GraphSAGE}} were obtained from their respective repositories. This approach ensured the use of standardized and well-maintained implementations for each baseline.

The experiments were conducted on six datasets from the Network Repository \cite{nr2015} and SNAP \cite{leskovec2016snap}, spanning diverse domains such as social, biological, and ecological networks. All graphs were treated as static by disregarding temporal information. Each graph was split into 70\% edges for training and 30\% for testing. Node embeddings were learned on the training graph, while the testing edges were excluded to ensure unbiased evaluation.

\begin{table}[h!]
\centering
\caption{Dataset Statistics}
\label{tab:dataset_statistics}
\resizebox{0.48\textwidth}{!}{%
\begin{tabular}{lccccc}
\toprule
\textbf{Dataset} & \(|V|\) & \(|E|\) & \(d_{\text{avg}}\) & \(K_{\text{avg}}\) & \(T_{\text{frac}}\) \\ 
\midrule
fb-pages-food      & 620   & 2091   & 6.7      & 0.33   & 0.22 \\ 
ego-Facebook       & 4039  & 88234  & 43.69    & 0.60   & 0.26 \\ 
soc-hamsterster    & 2000  & 16097  & 16.10    & 0.5375 & 0.2314 \\ 
aves-weaver-social & 117   & 304    & 5.20     & 0.6924 & 0.5747 \\ 
bio-DM-LC          & 483   & 997    & 4.13     & 0.1047 & 0.0551 \\ 
bio-WormNet-v3     & 2274  & 78328  & 68.89    & 0.8390 & 0.7211 \\ 
\bottomrule
\end{tabular}%
}
\begin{flushleft}
\footnotesize{\textbf{Notation:} 
\(|V|\): Number of nodes; 
\(|E|\): Number of edges; 
\(d_{\text{avg}}\): Average node degree; 
\(K_{\text{avg}}\): Average clustering coefficient; 
\(T_{\text{frac}}\): Fraction of closed triangles.
}
\end{flushleft}
\end{table}

The hyperparameters for CTW, DW, and N2V were set to \(L_w = 80\) (walk length), \(R = 10\) (number of walks per node), and \(d = 128\) (embedding dimension). Edge embeddings were computed using the Hadamard product of node embeddings. The performance was evaluated using logistic regression and two metrics: \textbf{AUC (Area Under the Curve)} and \textbf{Average Precision (AP)}. Each experiment was repeated ten times with different random seeds, and the average results are reported.

\subsection{Results and Analysis}
\paragraph{Link Prediction.}
The results of the link prediction task are presented in Table~\ref{tab:auc_static} (AUC) and Table~\ref{tab:ap_static} (AP). CTW consistently outperformed baseline methods across all datasets. Notably, CTW demonstrated significant improvements on datasets with strong community structures, such as \textit{soc-hamsterster} and \textit{fb-pages-food}. For instance:
\begin{itemize}
    \item On \textit{bio-WormNet-v3}, CTW achieved an AUC of 0.9915, outperforming the second-best method, M-NMF, by 3.2\%.
    \item On \textit{ego-Facebook}, CTW achieved an AP of 0.9891, showcasing robust performance in large-scale social networks.
    \item On datasets with smaller community structures, such as \textit{aves-weaver-social}, CTW achieved a notable improvement, with AUC gains of up to 5.1\% over the best baseline.
\end{itemize}

\begin{table*}[htbp]
    \centering
    \begin{minipage}{0.48\textwidth}
        \centering

        \resizebox{\textwidth}{!}{%
        \begin{tabular}{lccccccc}
        \toprule
        \textbf{Dataset} & \textbf{CTW} & \textbf{N2V} & \textbf{DW} & \textbf{LINE} & \textbf{S2V} & \textbf{GS} & \textbf{MN} \\ 
        \midrule
        fb-pages-food      & \textbf{0.9319} & 0.8764 & 0.8632 & 0.8121 & 0.8573 & 0.8902 & 0.8964 \\ 
        ego-Facebook       & \textbf{0.9885} & 0.9452 & 0.9203 & 0.8947 & 0.9345 & 0.9541 & 0.9721 \\ 
        soc-hamsterster    & \textbf{0.9122} & 0.8610 & 0.8423 & 0.8045 & 0.8534 & 0.8812 & 0.8734 \\ 
        aves-weaver-social & \textbf{0.9114} & 0.8732 & 0.8517 & 0.8245 & 0.8651 & 0.8832 & 0.8903 \\ 
        bio-DM-LC          & \textbf{0.9337} & 0.8810 & 0.8743 & 0.8547 & 0.8723 & 0.8901 & 0.9121 \\ 
        bio-WormNet-v3     & \textbf{0.9915} & 0.9543 & 0.9402 & 0.9134 & 0.9624 & 0.9742 & 0.9603 \\ 
        \bottomrule
        \end{tabular}%
        } 
           
                \caption{Average AUC Scores for Link Prediction on Static Graphs. Bold numbers represent the best results.}
                     \label{tab:auc_static}
    \end{minipage}
    \hfill
    \begin{minipage}{0.48\textwidth}
        \centering

        \resizebox{\textwidth}{!}{%
        \begin{tabular}{lccccccc}
        \toprule
        \textbf{Dataset} & \textbf{CTW} & \textbf{N2V} & \textbf{DW} & \textbf{LINE} & \textbf{S2V} & \textbf{GS} & \textbf{MN} \\ 
        \midrule
        fb-pages-food      & \textbf{0.9421} & 0.8920 & 0.8814 & 0.8289 & 0.8712 & 0.9045 & 0.9102 \\ 
        ego-Facebook       & \textbf{0.9891} & 0.9534 & 0.9351 & 0.9107 & 0.9443 & 0.9642 & 0.9813 \\ 
        soc-hamsterster    & \textbf{0.9215} & 0.8735 & 0.8602 & 0.8224 & 0.8698 & 0.8914 & 0.8803 \\ 
        aves-weaver-social & \textbf{0.9217} & 0.8801 & 0.8655 & 0.8371 & 0.8764 & 0.8942 & 0.9015 \\ 
        bio-DM-LC          & \textbf{0.9440} & 0.8954 & 0.8883 & 0.8651 & 0.8850 & 0.9025 & 0.9234 \\ 
        bio-WormNet-v3     & \textbf{0.9961} & 0.9602 & 0.9456 & 0.9175 & 0.9723 & 0.9815 & 0.9673 \\ 
        \bottomrule
        \end{tabular}%
        
        }
                \caption{Average AP Scores for Link Prediction on Static Graphs. Bold numbers represent the best results.}
                        \label{tab:ap_static}
    \end{minipage}
\end{table*}

\section{Theoretical Analysis}

We interpret CTWalks from the perspective of matrix factorization. We demonstrate that the two-layer random walk process in CTWalks corresponds to a novel matrix factorization form, integrating intra- and inter-community dynamics through a hierarchical transition mechanism. These analyses highlight CTWalks’ theoretical advantages in encoding community-aware structures, providing insights into its effectiveness in network representation learning.

\subsection*{Analysis 2: Matrix Factorization Perspective of CTWalks}

The Skip-Gram with negative sampling (SGNS) framework \cite{mikolov2013distributed} in network embedding methods such as DeepWalk \cite{perozzi2014deepwalk} and node2vec \cite{grover2016node2vec} has been shown to implicitly factorize a pointwise mutual information (PMI) matrix \cite{levy2014neural,Qiu:2018}. CTWalks  extends this perspective by incorporating a hierarchical, two-layer random walk that explicitly encodes both intra-community and inter-community transitions, resulting in a novel matrix factorization form. 

\subsubsection*{CTWalks Transition Matrices}

In CTWalk, the probability of transitioning between two nodes during the random walk is governed by two distinct components: \textit{intra-community transitions} (within communities) and \textit{inter-community transitions} (across communities). To model this, we introduce two matrices:  
1) a \textbf{block-diagonal matrix} \( M_I \in \mathbb{R}^{|V| \times |V|} \) that represents transitions within individual communities, and  
2) an \textbf{extended matrix} \( M_C  \in \mathbb{R}^{|V| \times |V|}\) that captures transitions between communities via bridging nodes.

\paragraph{Intra-community transition matrix (\( M_I \)).}  
The matrix \( M_I \) represents the normalized intra-community transitions, structured as a block-diagonal matrix. Each block \( D_i^{-1} A_i \) corresponds to a community \( C_i \), where:
\begin{itemize}
    \item \( A_i \in \mathbb{R}^{|V_i| \times |V_i|} \) is the adjacency matrix of the subgraph \( G_i = (V_i, E_i) \), induced by the set of nodes \( V_i \) and edges \( E_i \) within community \( C_i \).
    \item \( D_i^{-1} \) is the inverse degree matrix of \( A_i \), ensuring row normalization of transition probabilities within each community.
\end{itemize}

The resulting block-diagonal structure of \( M_I \) is expressed as:
\begin{equation}
M_I =
\begin{pmatrix}
D_1^{-1} A_1 & 0 & \cdots & 0 \\
0 & D_2^{-1} A_2 & \cdots & 0 \\
\vdots & \vdots & \ddots & \vdots \\
0 & 0 & \cdots & D_k^{-1} A_k
\end{pmatrix}.
\end{equation}

This formulation confines random walks within individual communities \( C_1, C_2, \dots, C_k \), effectively capturing the dense intra-community relationships while isolating transitions between distinct communities.

\paragraph{Inter-community transition matrix (\( M_C \)).}  
The matrix \( M_C \) captures transitions between inter-community bridging nodes. To construct \( M_C \), we first normalize the adjacency matrix \( A_c \) of \( G_c = (V_c, E_c) \), where \( V_c \) is the set of bridging nodes. The normalized matrix is defined as:
\begin{equation}
A_c^{(n)} = D_c^{-1} A_c,
\end{equation}
where \( D_c \) is the degree matrix of \( A_c \), ensuring that the rows of \( A_c^{(n)} \) sum to one. Here, \( A_c^{(n)} \in \mathbb{R}^{|V_c| \times |V_c|} \) represents the normalized transition probabilities restricted to the bridging nodes.

To integrate \( M_C \) with \( M_I \), we align the node order in \( M_C \) to match the node ordering used in \( M_I \). Specifically, the rows and columns of \( M_C \) are arranged such that:
\begin{itemize}
    \item Nodes within the same community are grouped together, maintaining the block-diagonal structure of \( M_I \),
    \item Rows and columns corresponding to non-bridging nodes \( w \notin V_c \) are filled with zeros.
\end{itemize}

This alignment ensures that both \( M_C \) and \( M_I \) are compatible for subsequent matrix operations, such as multi-step transition summation. The resulting \( M_C \) can be visualized as follows:

\begin{equation}
M_C =
\begin{pmatrix}
0 & 0 & 0 & 0 & 0 & \cdots & 0 \\
0 & A_c^{(n)}[v_2, v_2] & 0 & 0 & A_c^{(n)}[v_2, v_5] & \cdots & 0 \\
0 & 0 & 0 & 0 & 0 & \cdots & 0 \\
0 & 0 & 0 & 0 & 0 & \cdots & 0 \\
0 & A_c^{(n)}[v_5, v_2] & 0 & 0 & A_c^{(n)}[v_5, v_5] & \cdots & 0 \\
\vdots & \vdots & \vdots & \vdots & \vdots & \ddots & \vdots \\
0 & 0 & 0 & 0 & 0 & \cdots & 0
\end{pmatrix}.
\end{equation}

This design ensures that \( M_C \) and \( M_I \) remain consistent in terms of node alignment. By isolating inter-community transitions and aligning them to the global node order of \( M_I \), \( M_C \) facilitates a seamless integration of intra-community and inter-community dynamics in the CTWalks  framework.

\subsubsection*{Matrix Factorization of CTWalks}

\textbf{Lemma 2: } Let \( M_C \) and \( M_I \) represent the inter-community and intra-community transition matrices, respectively. The embeddings learned by CTWalks  using 
Skip-Gram with negative sampling correspond to a low-rank factorization of the following shifted Pointwise Mutual Information (PMI) matrix:
\begin{equation}
\log \left( \text{vol}(G) \cdot \left( \frac{1}{T} \sum_{r=1}^T \left( M_C^r + M_I^r \right) \right) D^{-1} \right) - \log k,
\end{equation}
where:
\begin{itemize}
    \item $\text{vol}(G)$ = $\sum_i \sum_j A_{i,j}$ is the volume of graph \( G \),
    \item \( T \) is the context window size,
    \item \( M_C^r \) and \( M_I^r \): The \( r \)-step transition probability matrices for \( M_C \) (inter-community transitions) and \( M_I \) (intra-community transitions), respectively,
    \item \( D^{-1} \) is the inverse degree matrix of \( G \),
    \item \( k \) is the negative sampling parameter.
\end{itemize}

\noindent\textbf{Proof.}  
We begin with the Skip-Gram objective, which seeks to maximize the probability of observing a context node \( c \) given a target node \( w \). Under the CTWalks  framework, random walks are independently applied to the inter-community transition matrix \( M_C \) and the intra-community transition matrix \( M_I \).

\paragraph{Step 1: Joint Probability.}  
The joint probability \( P(w, c) \) of observing a node \( w \) with context \( c \) is defined by the stationary distribution \( \pi(w) \) and the \( T \)-step transition probabilities:
\begin{equation}
P(w, c) \approx \pi(w) \cdot \frac{1}{T} \sum_{r=1}^T \left( M_C^r + M_I^r \right)[w, c],
\end{equation}
where \( \pi(w) \) is the stationary probability of node \( w \), given by:
\begin{equation}
     \pi(w) = \frac{d(w)}{\text{vol}(G)}.
\end{equation}

\paragraph{Step 2: Marginal Probabilities.}  
The marginal probability \( P(w) \) of node \( w \) under the stationary distribution is given by:
\begin{equation}
P(w) = \pi(w) = \frac{d(w)}{\text{vol}(G)}.
\end{equation}
Similarly, the marginal probability \( P(c) \) of context \( c \) is:
\begin{equation}
P(c) = \pi(c) = \frac{d(c)}{\text{vol}(G)}.
\end{equation}

\paragraph{Step 3: Pointwise Mutual Information (PMI).}  
The PMI between target node \( w \) and context node \( c \) is defined as:  
\begin{equation}
\label{PMI}
PMI(w, c) = \log \frac{P(w, c)}{P(w)P(c)}.
\end{equation}
Substituting \( P(w, c) \), \( P(w) \), and \( P(c) \) into Eq.(\ref{PMI}), we obtain:
\begin{equation}
PMI(w, c) \approx \log \frac{\pi(w) \cdot \frac{1}{T} \sum_{r=1}^T \left( M_C^r + M_I^r \right)[w, c]}{\pi(w) \pi(c)}.
\end{equation}
Simplifying:
\begin{equation}
PMI(w, c) \approx \log \left( \frac{1}{T} \sum_{r=1}^T \left( M_C^r + M_I^r \right)[w, c] \cdot \frac{\text{vol}(G)}{d(c)} \right).
\end{equation}

\paragraph{Step 4: Matrix Representation.}  
We can represent the PMI matrix in the following matrix form:
\begin{equation}
\log \left( \text{vol}(G) \cdot \left( \frac{1}{T} \sum_{r=1}^T \left( M_C^r + M_I^r \right) \right) D^{-1} \right).
\end{equation}

\paragraph{Step 5: SGNS Objective.}  
The Skip-Gram with negative sampling introduces a shift \( -\log k \), where \( k \) is the number of negative samples. Therefore, the final form of the factorization becomes:
\begin{equation}
\log \left( \text{vol}(G) \cdot \left( \frac{1}{T} \sum_{r=1}^T \left( M_C^r + M_I^r \right) D^{-1} \right) \right) - \log k.
\end{equation}

\textbf{Lemma 2} demonstrates that the embeddings learned by CTWalks  correspond to the low-rank factorization of the combined \( T \)-step transition probabilities derived from \( M_C \) and \( M_I \). This factorization effectively integrates inter-community and intra-community relationships, capturing both local and global relationships without requiring additional parameters.

\paragraph{Significance of the Factorization.}  
By combining \( M_C \) and \( M_I \) in a structured manner, CTWalks  enables an explicit encoding of both local community structures and global inter-community connectivity. This theoretical insight justifies the empirical superiority of CTWalks  in capturing hierarchical and multi-scale graph properties in tasks such as link prediction and node classification.

\section{Detailed Explanation of Node Anonymization}

\paragraph*{Motivation for Anonymization.}
The primary goal of anonymization is to achieve \textbf{generalization} across nodes, enabling the model to effectively operate on unseen nodes or structures during inference. By replacing explicit node identities with position-based representations:
\begin{itemize}
    \item The model avoids overfitting to specific node features.
    \item Community labels enrich the anonymized representation, embedding mesoscopic graph structures.
    \item Directionality-aware representations ensure that roles of nodes as sources or targets are preserved.
\end{itemize}

\paragraph*{Process Overview.}
Anonymization in CTWalks operates on temporal walks generated from a given interaction \(((u, v), t_i)\). For each interaction, two sets of walks are constructed:
\begin{itemize}
    \item \( \mathcal{S}_u \): Walks originating from the source node \( u \).
    \item \( \mathcal{S}_v \): Walks originating from the target node \( v \).
\end{itemize}
A unique set of nodes \( V_{\text{interaction}} \) is obtained from \( \mathcal{S}_u \cup \mathcal{S}_v \). Each node \( w \in V_{\text{interaction}} \) is anonymized by aggregating its positional occurrences across all walks in the respective sets, incorporating directionality and community context.

\paragraph*{Anonymization Steps.}
1. For each node \( w \) in \( V_{\text{interaction}} \), perform the following steps:
    \begin{itemize}
        \item \textbf{Position Vector Computation for \( \mathcal{S}_u \):}
        \begin{itemize}
            \item For each walk \( W \in \mathcal{S}_u \), compute \( w \)'s positional occurrence vector:
            \begin{equation}
            A(w; W) = [\mathbb{I}(w = W[1]), \mathbb{I}(w = W[2]), \dots, \mathbb{I}(w = W[l])],
            \end{equation}
            where \( l \) is the walk length and \( \mathbb{I} \) is an indicator function.
            \item Aggregate \( w \)'s positional vectors across all walks in \( \mathcal{S}_u \):
            \begin{equation}
            A(w; \mathcal{S}_u) = \sum_{W \in \mathcal{S}_u} A(w; W).
            \end{equation}
        \end{itemize}
        \item \textbf{Community Label Integration for \( \mathcal{S}_u \):}
        \begin{equation}
        A(w; \mathcal{S}_u, C_u) = [A(w; \mathcal{S}_u) || C_u],
        \end{equation}
        where \( C_u \) is the community label of the source node \( u \).
        \item \textbf{Repeat for \( \mathcal{S}_v \):}
        Similarly, compute:
        \begin{equation}
        A(w; \mathcal{S}_v, C_v) = [A(w; \mathcal{S}_v) || C_v],
        \end{equation}
        where \( C_v \) is the community label of the target node \( v \).
        \item \textbf{Combine Source and Target Representations:}
        Incorporate directionality by concatenating the source and target representations:
        \begin{equation}
        A(w; \mathcal{S}_u, \mathcal{S}_v, C_u, C_v) = [A(w; \mathcal{S}_u, C_u) || A(w; \mathcal{S}_v, C_v)].
        \end{equation}
    \end{itemize}

2. Store the final anonymized representation \( A(w; \mathcal{S}_u, \mathcal{S}_v, C_u, C_v) \) for all \( w \in V_{\text{interaction}} \).

\paragraph*{Anonymized Walk Construction.}
After anonymizing individual nodes, the anonymized temporal walk is constructed. For a single walk \( W = \{w_1, w_2, \dots, w_l\} \) with timestamps \( t_1 < t_2 < \dots < t_l \), the anonymized walk is defined as:
\begin{equation}
W_{\text{anon}} = \{(A(w_1), t_1), (A(w_2), t_2), \dots, (A(w_l), t_l)\},
\end{equation}
where \( A(w_i) \) is the direction- and community-aware anonymized representation of \( w_i \).

\paragraph*{Discussion.}
This anonymization process achieves the following:
\begin{itemize}
    \item \textbf{Generalization:} By removing raw node identities and replacing them with position-based representations, the process ensures generalization across different nodes and graph structures.
    \item \textbf{Directionality:} Separate representations for \( \mathcal{S}_u \) and \( \mathcal{S}_v \) preserve the source and target node roles, capturing directional interactions.
    \item \textbf{Community Awareness:} Incorporating community labels \( C_u \) and \( C_v \) enhances context-awareness, allowing the model to leverage mesoscopic structural information.
\end{itemize}
This integration of positional, directional, and community-based information establishes a strong foundation for robust representation learning in CTWalks. The anonymization process is detailed in Algorithm \ref{alg:anonymization}.

\begin{algorithm}[H]
\caption{Node Anonymization in CTWalks for a Given Interaction}
\label{alg:anonymization}
\textbf{Input:} Temporal interaction \(((u, v), t_i)\), walk sets \( \mathcal{S}_u \) and \( \mathcal{S}_v \), community labels \( C_u \) and \( C_v \). \\
\textbf{Output:} Anonymized representations \( \{A(w; \mathcal{S}_u, \mathcal{S}_v, C_u, C_v) \mid w \in V_{\text{interaction}} \} \), where \( V_{\text{interaction}} \) is the set of unique nodes appearing in \( \mathcal{S}_u \) and \( \mathcal{S}_v \).

\begin{algorithmic}[1]
\STATE Initialize \( V_{\text{interaction}} \gets \text{unique nodes in } \mathcal{S}_u \cup \mathcal{S}_v \).
\STATE Initialize \( \text{AnonymizedList} \gets \emptyset \).

\FOR{each node \( w \in V_{\text{interaction}} \)}
    \STATE Initialize \( A(w; \mathcal{S}_u) \gets \mathbf{0} \) (zero vector of length equal to the walk length).
    \FOR{each walk \( W \in \mathcal{S}_u \)}
        \STATE Compute position vector \( A(w; W) = [\mathbb{I}(w = W[1]), \dots, \mathbb{I}(w = W[l])] \).
        \STATE Aggregate position vectors: \( A(w; \mathcal{S}_u) \gets A(w; \mathcal{S}_u) + A(w; W) \).
    \ENDFOR
    \STATE Append community label: \( A(w; \mathcal{S}_u, C_u) \gets [A(w; \mathcal{S}_u) || C_u] \).

    \STATE Initialize \( A(w; \mathcal{S}_v) \gets \mathbf{0} \) (zero vector of length equal to the walk length).
    \FOR{each walk \( W \in \mathcal{S}_v \)}
        \STATE Compute position vector \( A(w; W) = [\mathbb{I}(w = W[1]), \dots, \mathbb{I}(w = W[l])] \).
        \STATE Aggregate position vectors: \( A(w; \mathcal{S}_v) \gets A(w; \mathcal{S}_v) + A(w; W) \).
    \ENDFOR
    \STATE Append community label: \( A(w; \mathcal{S}_v, C_v) \gets [A(w; \mathcal{S}_v) || C_v] \).

    \STATE Concatenate direction-aware representations:
    \begin{equation}
    A(w; \mathcal{S}_u, \mathcal{S}_v, C_u, C_v) \gets [A(w; \mathcal{S}_u, C_u) || A(w; \mathcal{S}_v, C_v)].
    \end{equation}
    \STATE Add \( A(w; \mathcal{S}_u, \mathcal{S}_v, C_u, C_v) \) to \( \text{AnonymizedList} \).
\ENDFOR

\STATE \textbf{Return} \( \text{AnonymizedList} \).
\end{algorithmic}
\end{algorithm}

\section{Comparison and Advantages of Continuous-Time Approaches in CTWalks}

Continuous-time models have gained significant traction in recent years, offering an alternative to traditional discrete-time approaches for modeling irregularly sampled time series and graph data. Below, we explore different continuous-time methodologies and position CTWalks within this landscape, demonstrating its unique contributions and advantages.

\begin{figure}[h!]
    \centering
    \includegraphics[width=0.4\textwidth]{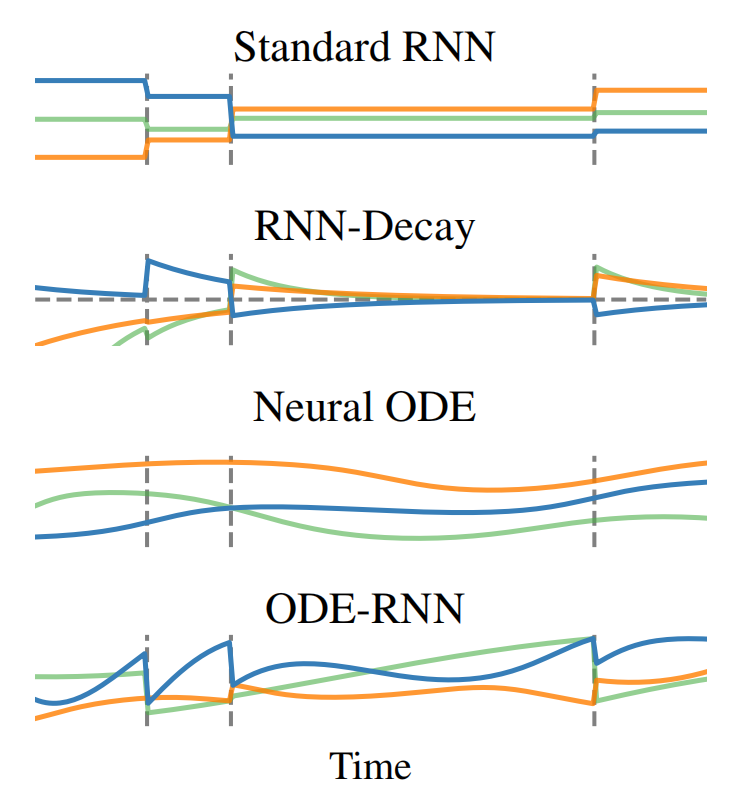}
    \caption{Comparison of temporal dynamics modeling approaches. Standard RNN maintains constant hidden states; RNN-Decay introduces exponential decay; Neural ODE models smooth continuous trajectories; ODE-RNN combines discrete updates with continuous dynamics.}
    \label{fig:continuous_methods}
\end{figure}

\subsection*{Overview of Continuous-Time Approaches}

\textbf{Standard RNNs.} Recurrent Neural Networks (RNNs) excel at modeling high-dimensional, regularly sampled time series such as text and speech. However, they struggle with irregular sampling, as they inherently assume fixed time intervals between observations. This limitation necessitates preprocessing steps such as imputation or aggregation, which can lead to loss of critical temporal information~\cite{lipton2016critical, che2018recurrent}.

\textbf{RNN-Decay.} To address the irregularity issue, RNN-Decay introduces exponentially decaying hidden states between observations. While this approach maintains some temporal dynamics, it assumes a simplistic decay function, which limits its ability to capture complex temporal dependencies~\cite{mei2017neural}.

\textbf{Neural ODEs.} Neural Ordinary Differential Equations (ODEs)~\cite{chen2018neural} define hidden state dynamics as continuous functions governed by ODEs, offering a principled way to model time as a continuous variable. Neural ODEs eliminate the need for fixed time steps, making them well-suited for irregularly sampled data. However, their reliance on purely continuous trajectories can oversimplify scenarios involving abrupt changes or discrete transitions.

\textbf{ODE-RNNs.} ODE-RNNs~\cite{rubanova2019latent} extend Neural ODEs by combining continuous dynamics with discrete updates. Between observations, hidden states evolve via ODE solvers, while at observation points, updates are applied using neural networks. This hybrid approach captures both continuous and discrete aspects of temporal data, making it particularly effective for sparse time series.

Figure~\ref{fig:continuous_methods} illustrates continuous-time approaches introduce enhancements for dynamic graphs. Standard RNNs fail to model irregular intervals effectively, while RNN-Decay offers limited improvements. Neural ODEs provide continuous trajectories but lack discrete adaptability. ODE-RNNs bridge this gap, and CTWalks extends this framework further by embedding community context into temporal dynamics. Building on the strengths of ODE-RNNs, CTWalks incorporates continuous-time dynamics into graph representation learning. Unlike prior methods, CTWalks integrates a community-aware framework with ODE-driven sampling, offering the following advantages:

\begin{itemize}
    \item \textbf{Preservation of Temporal Dynamics:} By employing ODE solvers to model temporal transitions, CTWalks accurately captures both intra- and inter-community dynamics without requiring imputation or aggregation. This aligns with the principles of Neural ODEs but extends their applicability to dynamic graphs.
    
    \item \textbf{Hybrid Continuous-Discrete Framework:} Similar to ODE-RNNs, CTWalks combines continuous-time evolution with discrete updates. This enables the model to adapt to abrupt changes in graph structure, such as the formation or dissolution of communities.
    
    \item \textbf{Community-Aware Sampling:} The integration of community-aware mechanisms ensures that temporal walks respect community boundaries while capturing cross-community interactions. This is a key distinction from traditional ODE-based methods, which lack structural awareness.
\end{itemize}

By embedding continuous-time dynamics into a community-aware framework, CTWalks bridges temporal modeling with structural awareness, enabling more sophisticated applications in dynamic graph representation learning.

\end{document}